\documentclass{article}

\usepackage[preprint]{neurips_2021}

\usepackage[utf8]{inputenc} 
\usepackage[T1]{fontenc}    
\usepackage{hyperref}       
\usepackage{url}            
\usepackage{booktabs}       
\usepackage{amsfonts}       
\usepackage{nicefrac}       
\usepackage{microtype}      
\usepackage{xcolor}         

\usepackage{subfig}
\usepackage{graphicx}

\usepackage{amsmath}
\usepackage{amssymb}
\usepackage{caption}
\usepackage{bm}
\usepackage{wrapfig}
\usepackage{sidecap}

\usepackage{algorithm}
\usepackage{algorithmic}

\title{Meta Automatic Curriculum Learning}

\author{
    R\'emy Portelas \\
    INRIA\\
    France\\
    \footnotesize{\texttt{remy.portelas@inria.fr}}\\
  \And
    Cl\'ement Romac \\
    INRIA\\
    France\\
 \And
    Katja Hofmann \\
    Microsoft Research\\
    Cambridge\\
  \And
    Pierre-Yves Oudeyer \\
    INRIA\\
    France\\
  }

\begin{document}

\maketitle

\begin{abstract}
A major challenge in the Deep RL (DRL) community is to train agents able to generalize their control policy over situations never seen in training. Training on diverse tasks has been identified as a key ingredient for good generalization, which pushed researchers towards using rich procedural task generation systems controlled through complex continuous parameter spaces. In such complex task spaces, it is essential to rely on some form of Automatic Curriculum Learning (ACL) to adapt the task sampling distribution to a given learning agent, instead of randomly sampling tasks, as many could end up being either trivial or unfeasible. Since it is hard to get prior knowledge on such task spaces, many ACL algorithms explore the task space to detect progress niches over time, a costly tabula-rasa process that needs to be performed for each new learning agents, although they might have similarities in their capabilities profiles. To address this limitation, we introduce the concept of Meta-ACL, and formalize it in the context of black-box RL learners, i.e. algorithms seeking to generalize curriculum generation to an (unknown) distribution of learners. In this work, we present AGAIN, a first instantiation of Meta-ACL, and showcase its benefits for curriculum generation over classical ACL in multiple simulated environments including procedurally generated parkour environments with learners of varying morphologies. Videos and code are available at {\footnotesize\url{https://sites.google.com/view/meta-acl}}.
\end{abstract}

\section{Introduction}

The idea of organizing the learning sequence of a machine is an old concept that stems from multiple works in reinforcement learning \citep{selfridge,Schmid}, developmental robotics \citep{oudeyer2007intrinsic} and supervised learning \citep{elman, bengiocl}, from which the Deep RL community borrowed the term \textit{Curriculum Learning}. Automatic CL \citep{portelas2020-acl-drl} refers to \textit{teacher} algorithms able to autonomously adapt their task sampling distribution to their evolving \textit{student}. In DRL, ACL has been leveraged to scaffold learners in a variety of multi-task control problems, including video-games \citep{icm,rnd,montezuma-single-demo}, multi-goal robotic arm manipulation \citep{her,curious,cideron2019self,fournier-accuracy-acl} and navigation in sets of environments \citep{tscl,portelas2019,ADRmila,goalgan,selfpaceddrl}. Concurrently, multiple authors demonstrated the benefits of Procedural Content Generation (PCG) as a tool to create rich task spaces to train generalist agents \citep{risiPCG,illuminating,quantif-coinrun}. The current limit of ACL is that, when applied to such large continuous task spaces, that often have few learnable subspaces, it either relies on 1) human expert knowledge that is hard/costly to provide (and which undermines how automatic the ACL approach is), or 2) it loses a lot of time finding tasks of appropriate difficulty through \textit{task exploration}.

Given the aforementioned impressive results on training DRL learners with ACL to generalize over tasks (which extended the classical single-task scenarios \citep{dqn,trpo,ddpg} to multi-tasks), we propose to go further and work on training (unknown) distributions of students on continuous task spaces, thereafter referred to as \textit{Classroom Teaching} (CT). CT defines a family of problems in which a teacher algorithm is tasked to sequentially generate multiple curricula tailored for each of its students, all having potentially varying abilities. CT differs from the problems studied in population-based developmental robotics \citep{imgep} and evolutionary algorithms \citep{poet} as in CT there is no direct control over the characteristics of learners, and the objective is to foster maximal learning progress over all learners rather than iteratively populating a pool of high-performing task-expert policies. Studying CT scenarios brings DRL closer to assisted education research problems and might stimulate the design of methods that alleviate the expensive use of expert knowledge in current state of the art methods \citep{zpdes, Koedinger13}. CT can also be transposed to (multi-task) robotic training scenarios, e.g. when performing iterative design improvements on a robot, which requires to train a sequence of morphologically related (yet different) robots.


Given multiple students to train, no expert knowledge, and assuming at least partial similarities between each students' optimal curriculum, current \textit{tabula-rasa} exploratory-ACL approaches that do not reuse knowledge between different students do not seem like the optimal choice. This motivates the research of what we propose to call Meta Automatic Curriculum Learning mechanisms, that is algorithms learning to generalize ACL over multiple students.
In this work we formalize this novel setup and propose a first Meta-ACL baseline algorithm (based on an existing ACL method \citep{portelas2019}). Given a new student to train, our approach is centered on the extraction of adapted curriculum priors from a history of previously trained students. The prior selection is performed by matching competence vectors that are built for each student through pre-testing. We show that this simple method can bring significant performance improvements over classical ACL in both a toy environment without DRL students and on Box2D parkour environments with DRL learners.

\paragraph{Related Work.} To approach the problem of curriculum generation for DRL agents, recent works proposed multiple ACL algorithms based on the optimization of surrogate objectives such as learning progress \citep{portelas2019,tscl,tscllike,curious}, diversity \citep{diayn,metarl-carml,countbased} or intermediate difficulty \citep{goalgan,settersolver,OpenAI2019SolvingRC,reverse-cur,ADRmila}. All these works tackled student training through independent ACL runs, while we propose to investigate how one can share information accross multiple trainings.
Within DRL, \textit{Policy Distillation} \citep{distralTeh2017,pol-dil-review} consists in leveraging one or several previously trained policies to perform \textit{behavior cloning} on a new policy (e.g. to speed up training and/or to leverage task-experts to train a multi-task policy). Our work can be seen as proposing a complementary toolbox aiming to perform \textit{Curriculum Distillation} on a continuous space of tasks.

Similar ideas were developed for supervised learning by \citep{hacohen19a-scoring-pacing, ban, banlike}. In \citep{hacohen19a-scoring-pacing}, authors propose an approach to infer a curriculum from past training for an image classification task: they train their network once without curriculum and use its predictive confidence for each image as a difficulty measure exploited to derive an appropriate curriculum to re-train the network. Although we are mainly interested in training a classroom of diverse students, section \ref{exp:trying-again} presents similar experiments in a DRL scenario, showing that our Meta-ACL procedure can be beneficial for a single learner that we train once and re-train using curriculum priors inferred from the first run.

Parallel to this work, Turcheta et. al. \citep{safe-rl-curr-induction} studied how to infer safety constraints (i.e. curriculum) over multiple DRL students in a data-driven way to better perform on a given single task. In their work, students are only varying by their network's initialization and their teacher assumes the existence of a pre-defined discrete set of safety constraints to choose from. By contrast, we consider the problem of training generalist students with varying morphologies (and networks initializations), with a teacher algorithm choosing tasks from a continuous task space.



\paragraph{Main Contributions.}
\begin{itemize}
    \item Introduction to the concept of Meta-ACL, i.e. algorithms that generalize curriculum generation in Classroom Teaching scenarios, and an approach to study these algorithms. Formalization of the interaction flows between Meta-ACL algorithms and (unknown) Deep RL student distributions.
    \item Introduction of AGAIN, a first Meta-ACL baseline algorithm which learns curriculum priors to fasten the identification of learning progress niches for new Deep RL students.
    \item Design of a toy-environment and of a parametric Box2D Parkour environment featuring a multi-modal distribution of possible agent embodiments well suited to study Meta-ACL.
    \item Analysis of AGAIN on these environments, demonstrating the performance advantages of this approach over classical ACL, including (and surprisingly) when applied to a single student.
  
\end{itemize}

\section{Meta Automatic Curriculum Learning Framework}
\label{sec:framework}

\begin{figure}[htb!]
\centering
\includegraphics[width=\textwidth]{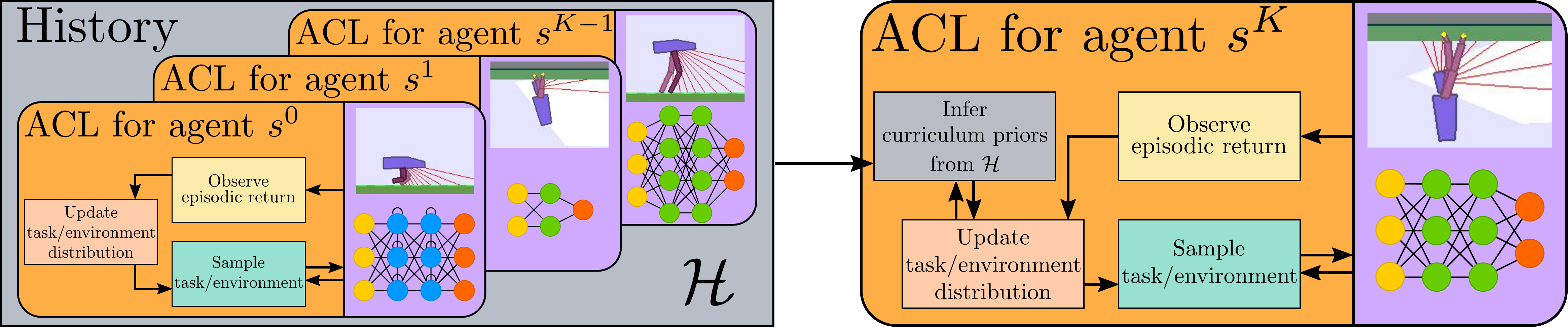}
\caption{\footnotesize{In \textit{Meta Automatic Curriculum Learning }(Meta-ACL), the objective is to leverage previous teaching experience to improve the curriculum generation of a new agent, whose embodiment and learning mechanisms have potentially never been seen before: the teacher has to \textit{generalize} over students.}}
\label{meta-acl-fig}
\end{figure}

\paragraph{Black-box students} The Meta-ACL framework assumes the existence of policy learners, a.k.a students, capable of interacting in episodic control tasks. These students are assumed non-resettable, as in classical ACL scenarios. Their optimization objective is the maximization of some performance measure $P$ w.r.t the task (e.g. episodic reward, exploration). To make the framework problem-independant, we do not assume expert knowledge over the task space w.r.t the student distribution, e.g. task subspaces could be trivial for some students and unfeasible for others. The objective of Meta-ACL is precisely to autonomously infer such prior knowledge from experience in scenarios where human expert knowledge is either hard or impossible to use. Similarly, we consider a black-box teaching scenario, i.e. we do not assume the knowledge of which learning mechanisms are used by students (e.g. DRL agents, evolutionary algorithms, ...).


\paragraph{Automatic Curriculum Learning} \hspace{-0.2cm} Given such a black-box student $s$ to train on a continuous \textit{task space} $\mathcal{A}$, the purpose of an ACL algorithm is to sequentially sample (parameterized) tasks for $s$, such that the following evaluation metric, used by the experimenter, is maximized:
\begin{equation}
    \label{eq:acl-obj}
    \max \int_{a\sim \mathcal{A}} \! P_{s,a}^E\, \mathrm{d}a,
\end{equation}
with $E$ the episode budget, and $P_{s,a}^E$ the end performance of student $s$ on task $a$ (e.g. exploration score, cumulative reward). Since direct optimization of such a post-training performance is difficult, ACL is often approached using proxy objectives (e.g. intermediate difficulty). Given one such proxy objective, an ACL algorithm usually relies on observing episodic behavioral features of its student w.r.t to proposed tasks (e.g. episodic rewards), allowing to infer a competence status $o \in \mathcal{O}$ of its learner (e.g. progress regions in $\mathcal{A}$) which conditions task sampling. This sequential task selection unrolled by ACL methods along a student's training can be transposed into a policy search problem on a high-level non-episodic POMDP. While interacting within this POMDP, an ACL policy $\Pi(o,s) \rightarrow \rho(\mathcal{A})$ proposes task distributions $\rho(\mathcal{A}) \subset \mathcal{A}$ to its student, observes the resulting behavioral features to update its competence status $o$, and collects objective-dependant rewards (e.g. learning progress).


In practice, approaching this task-level control problem with classical DRL algorithms is challenging because of sample efficiency: an ACL policy has to be learned and exploited along interaction windows typically around a few tens of thousands of steps. This has to be compared to the tens of millions or sometimes billions of interaction steps necessary to train a DRL policy for robotic control tasks. For this reason, most recent ACL research has focused on reducing the teaching problem into a Multi Armed Bandit setup, which ignores the sequential dependency over student states implied in POMDP settings \citep{tscl,tscllike,curious,portelas2019}. Although out of the scope of this paper, the use of classical DRL approaches as Meta-ACL algorithms is worth investigating in future work.

\paragraph{Meta-ACL for Classroom Teaching.} We now present the concept of Meta-ACL applied to a Classroom Teaching scenario, i.e. there is no longer a single student $s$ to be trained, but a set of students with varying abilities (e.g. due to morphology and/or learning mechanisms) sequentially drawn from an unknown distribution $\mathcal{S}$. The notion of meta-learning refers to \textit{any type of learning guided by prior experience with other
tasks} \citep{surveymetarl}. In meta-RL, agents are \textit{learning to learn} to act \citep{wang2016}, i.e. their objective is to maximize performance on previously unseen test tasks after $0$ or a few learning updates. In other words, it is about leveraging knowledge from previously encountered tasks to generalize on new tasks. We propose to extend this concept into Meta-ACL, that is, algorithms that are \textit{learning to learn} to teach, i.e. they leverage knowledge from curricula built for previous students to improve the curriculum generation for new ones.  More precisely, a Meta-ACL algorithm can be formulated as a function:
\begin{equation} \label{eq:meta-acl}
\begin{gathered}
f(\Pi, \mathcal{H}, s^{K}) \rightarrow \Pi^{'} ~~s.t.~~\mathcal{H}=[\tau_{s^0}, \tau_{s^1}, ...,
\tau_{s^{K-1}}] \\
 \tau_{s^{i}} = [(\rho^0(\mathcal{A}), o^0), (\rho^1(\mathcal{A}), o^1), ..., (\rho^T(\mathcal{A}), o^{T})],
\end{gathered}
\end{equation}
with $\mathcal{H}$ the history of past $K$ training trajectories $\tau_s$ resulting from the scaffolding of $K$ previous students with an ACL (or Meta-ACL) policy $\Pi$, and $s^K$ the current student. Given our formalization of ACL (see eq. \ref{eq:acl-obj}), the experimenter's evaluation objective for Meta-ACL can be expressed as follows:
\begin{equation}
    \label{eq:meta-acl-obj}
    \max_{f} \int_{s\sim \mathcal{S}}\int_{a\sim \mathcal{A}} \! P_{s,a}^E\, \mathrm{d}a~\mathrm{d}s.
\end{equation}

As in the case of the ACL evaluation objective expressed in eq. \ref{eq:acl-obj}, direct optimization of eq. \ref{eq:meta-acl-obj} is difficult as it implies the joint maximization of multiple students' performance. In our experiments, we reduce the Meta-ACL problem to the sequential independent training of a set of new students by leveraging priors from previous student trainings (with the hope to maximize performance over the entire set). Figure \ref{meta-acl-fig} provides a visual transcription of the workflow of a Meta-ACL algorithm. While in our experiments we use a fixed-size history $\mathcal{H}$ of ACL-trained students (to make our experiments computationally tractable), $\mathcal{H}$ could be grown incrementally by collecting training trajectories online from Meta-ACL trainings.


\section{A first Meta-ACL baseline: Alp-Gmm And Inferred Progress Niches}
\label{sec:methods}

\begin{figure*}[b]
\centering
\includegraphics[width=\textwidth]{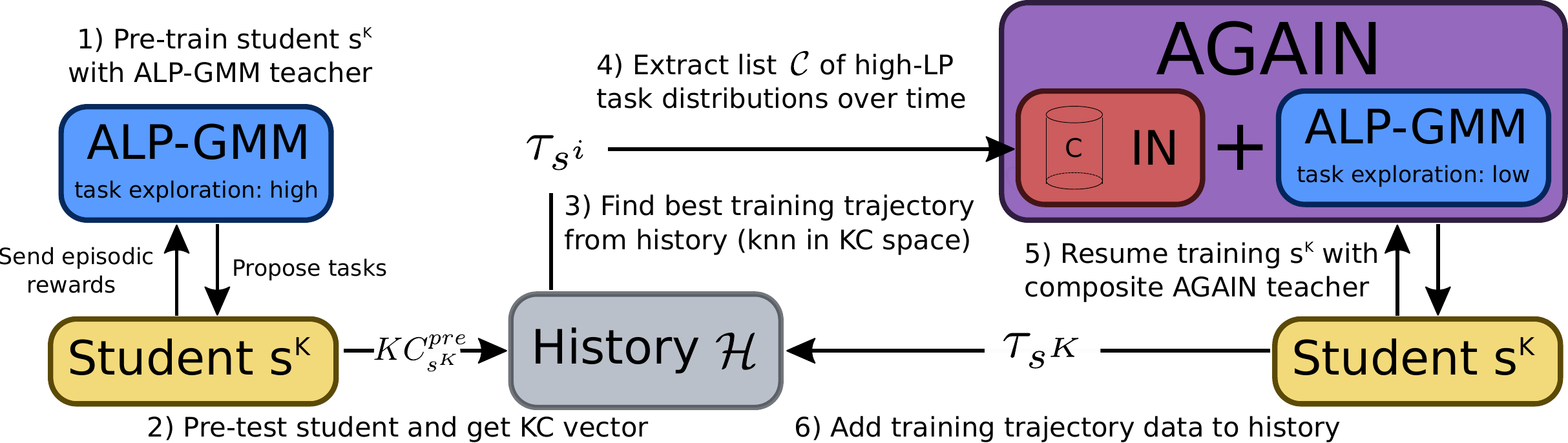}
\caption{\footnotesize{Schematic pipeline of Alp-Gmm And Inferred progress Niches (AGAIN), our proposed approach, which first leverages a preliminary run with a high-exploration ALP-GMM curriculum generator, and uses student pre-testing to infer an expert curriculum combined with a low-exploration ALP-GMM.}}
\label{again-fig}
\end{figure*}

In this section, we present AGAIN (Alp-Gmm And Inferred Progress Niches), our proposed Meta-ACL algorithm, and connect it to the formalism described in section \ref{sec:framework}. We first give a broad overview of the approach and then provide detailed explanations of key components.

\paragraph{Overview} Figure \ref{again-fig} provides a schematic pipeline of our Meta-ACL approach. Given a history $\mathcal{H}$ of previously trained students and a new student $s^K \sim \mathcal{S}$ to train, AGAIN starts by (1) pre-training $s^K$ using ALP-GMM, an existing ACL algorithm from \cite{portelas2019} (chosen for its simplicity and its non-reliance on expert knowledge). After pre-training, it (2) challenges the student with a set of test tasks to construct a meaningful competence profile of the student, which we thereafter refer to as a \textit{Knowledge Component (KC) vector}. This KC vector is then (3) used to select a previously trained student $s^i$ similar to $s^K$ from which the \textit{training trajectory} $\tau_{s^{i}}$ is recovered. Based on $\tau_{s^{i}}$, (4) AGAIN infers a set of curriculum priors $\mathcal{C}$ (i.e. promising task sub-spaces). Finally, (5) the training of $s^K$ can resume using a composite curriculum generator using both an expert curriculum derived from $\mathcal{C}$ (for exploitation), and ALP-GMM (for exploration).


\paragraph{1 - ALP-GMM} ALP-GMM \citep{portelas2019} is a Learning Progress (LP) based ACL technique for continuous task spaces that does not assume prior knowledge. ALP-GMM frames the task sampling problem into a non-stationary Multi-Armed bandit setup \citep{auer2002nonstochastic} in which arms are Gaussians spanning over the task space. The utility of each Gaussian is defined with a local LP measure derived from episodic reward comparisons. The essence of ALP-GMM is to periodically fit a Gaussian Mixture Model (GMM) on recently sampled tasks' parameters \textit{concatenated with their respective LP}. This periodically updated GMM used for task sampling can be seen as the evolving competence status $o$ described in section \ref{sec:framework}. The Gaussian from which to sample a new task is chosen proportionally to its mean LP dimension. Task exploration happens initially through a bootstrapping period of random task sampling and during training through residual random task sampling.

\paragraph{2,3 - KC-based curriculum priors selection.} For a new student $s^K$, given its capabilities on the considered task space, how to selected the most relevant previously trained student from which to extract curriculum priors? This problem is closely related to knowledge assessment in Intelligent Tutoring Systems setups studied in the educational data mining literature \citep{vie-mooc-test-assembly,vie-thesis}. Inspired by these works, we use pre-tests to derive a Knowledge Component vectors $KC^{pre} \in \mathbb{R}^{m}$ for all trained students. Each dimensions of $KC^{pre}$ contains the episodic return of the student on the corresponding pre-test task. Given that we do not assume access to expert knowledge, we build this pre-test task set by selecting $m$ tasks uniformly over the task space. We use the same task set to build a post-training KC vector $KC^{post} \in \mathbb{R}^{m}$ whose dimensions are summed up to get a score $j_s \in \mathbb{R}$, used to evaluate the end performance of students in $\mathcal{H}$. After the initial pre-training of $s^K$ with ALP-GMM, curriculum priors can be obtained in 3 steps: 1) pre-test $s^K$ to get its KC vector $KC^{pre}_{s^K}$, 2) infer the $k$ most similar previously trained students in KC space (using a k-nearest neighbor algorithm), and 3) use the training trajectory $\tau_s$ of the student with maximal post-training score $j_s$ among those $k$. In essence, this method is about re-using curriculum data from a similarly-skilled and successfully-trained student.


\paragraph{4 - Inferred progress Niches (IN).} Given that the KC-based student selection identified the training trajectory $\tau_{s^{i}}$ as the most promising for $s^K$, and assuming ALP-GMM as the underlying ACL teacher used for $s^i$, we can derive an expert curriculum from $\tau_{s^{i}}$ by first considering the ordered sequence of GMMs $\mathcal{C}_{raw}$ that were periodically fitted along training:

\begin{equation}  \label{eq:craw}
  \begin{gathered}
 \mathcal{C}_{raw} = \{p(1), ..., p(T)\} \\
 s.t.~~~p(t)= \sum_{i=1} LP_{ti}\mathcal{N}(\bm{\mu_{ti}},\bm{\Sigma_{ti}}),
\end{gathered}  
\end{equation}

with $T$ the total number of GMMs in the list and $LP_{ti}$ the Learning Progress of the $i^{th}$ Gaussian from the $t^{th}$ GMM. By keeping only Gaussians with $LP_{ti}$ above a predefined threshold $\delta_{LP}$, we can get a curated list $\mathcal{C}$ containing only Gaussians located on task subspaces on which $s^i$ experienced learning progress (i.e. curriculum priors). Given a GMM of $\mathcal{C}$, a task is selected by 1) sampling a Gaussian proportionally to its $LP_{ti}$ value, and 2) sampling the Gaussian to obtain parameters mapping to a task. But how to decide which GMMs to use along the training of the new student $s^K$ ?

While the simplest way to obtain such a curriculum would be to start sampling tasks from the first GMM and step to the next GMM at the same rate than the initial ALP-GMM run, we propose a more flexible \textit{reward-based} method. This method requires to record the mean episodic reward obtained by the previously trained student $s^i$ for each GMM of $\mathcal{C}$ (which can be done without additional assumptions or computational overhead). Given this, to select which GMM from $\mathcal{C}$ is used to sample tasks over time along the training of $s^K$, we start with the first GMM and only iterate over $\mathcal{C}$ once the mean episodic reward over tasks recently sampled from the current GMM matches or surpasses the mean episodic reward recorded during the initial ALP-GMM run. In app. \ref{ann:toy-exp}, we show that this \textit{reward-based} variant outperforms other potential methods. See app. \ref{app-again} for algorithmic details. We name the resulting meta-learned expert curriculum approach Infered progress Niches (IN)

\paragraph{5 - Our proposed approach: AGAIN.} Simply using IN directly for $s^K$ lacks adaptive mechanisms towards the characteristics of the new student (e.g. new embodiment, different initial parameters, ...), which could lead to failure cases where the expert curriculum misses important aspects of training (e.g. detecting task subspaces that are being forgotten). Additionally, the meta-learned ACL algorithm must have the capacity to emancipate from the expert curriculum once the trajectory is completed (i.e. go beyond $\mathcal{C}$). This motivates why our approach combines IN with an ALP-GMM teacher after the initial pre-training. The resulting Alp-Gmm And Inferred progress Niches approach (AGAIN) samples tasks from a GMM that is composed of the current mixture of both ALP-GMM and IN. See appendix \ref{app-alp-gmm} \& \ref{app-again} for implementation details and pseudo-code algorithms.

\section{Experiments and Results}
\label{sec:expes}

We organize the analysis of our proposed Meta-ACL algorithm around $3$ experimental questions:
\begin{itemize}
    \item What are the properties and important components of AGAIN? In this section we will leverage a toy environment without DRL students to conduct systematic experiments.
    \item Does AGAIN scale well to Meta-ACL scenarios with DRL students? Here we will present a new Parkour environment that will be used to conduct our experiments.
    \item Can AGAIN be used for single learners? Here we will show that it can be useful to derive curriculum priors even for a single student (i.e. without any student History $\mathcal{H}$).
\end{itemize}

\paragraph{Considered baselines and AGAIN variants.} In the following experiments we compare AGAIN to variants where 1) we directly use the expert curriculum instead of combining it with ALP-GMM (\textit{IN} condition), and 2) where we select a training trajectory at random (\textit{AGAIN\_RND}) or using the ground truth student distribution (\textit{AGAIN\_GT}). We compare these Meta-ACL variants to ACL approaches such as random curriculum generation (\textit{Random}), \textit{ALP-GMM} and either Adaptive Domain Randomization (\textit{ADR}) \citep{OpenAI2019SolvingRC} or an expert-made \textit{Oracle} curriculum. See appendix \ref{an:details} for details.

\subsection{Analysing Meta-ACL in a toy environment}
\label{sec:exp:toy-env}

To provide in-depth experiments on AGAIN, we first emancipate from DRL students through the use of a modified version of the toy testbed presented in \citep{portelas2019}. The objective of this environment is to simulate the learning of a student within a $2$D parameter space $\mathcal{P} = [0,1]^2$. The parameter space is uniformly divided in $400$ square cells $C \subset  \mathcal{P}$, and each parameter $p \in \mathcal{P}$ sampled by the teacher is directly mapped to an episodic reward $r_p$ based on sampling history and whether $C$ is considered "locked" or "unlocked". Three rules enforce reward collection in $\mathcal{P}$:~1) Every cell $C$ starts "locked", except a randomly chosen one that is "unlocked". 2) If $C$ is "unlocked" and $p \in C$, then $r_p = min(|C|,100)$, with $|C|$ the cumulative number of parameters sampled within $C$ while being "unlocked" (if $C$ is "locked", then $r_p = 0$). Finally, 3) If $|C| >= 75$, adjacent cells become "unlocked". Given these rules, one can model students with different curriculum needs by assigning them different initially unlocked cells, which itself models what is "easy to learn" initially for a given student, and from where it can expand.
\begin{figure*}[htb!]
\centering
\subfloat{\includegraphics[width=0.45\textwidth]{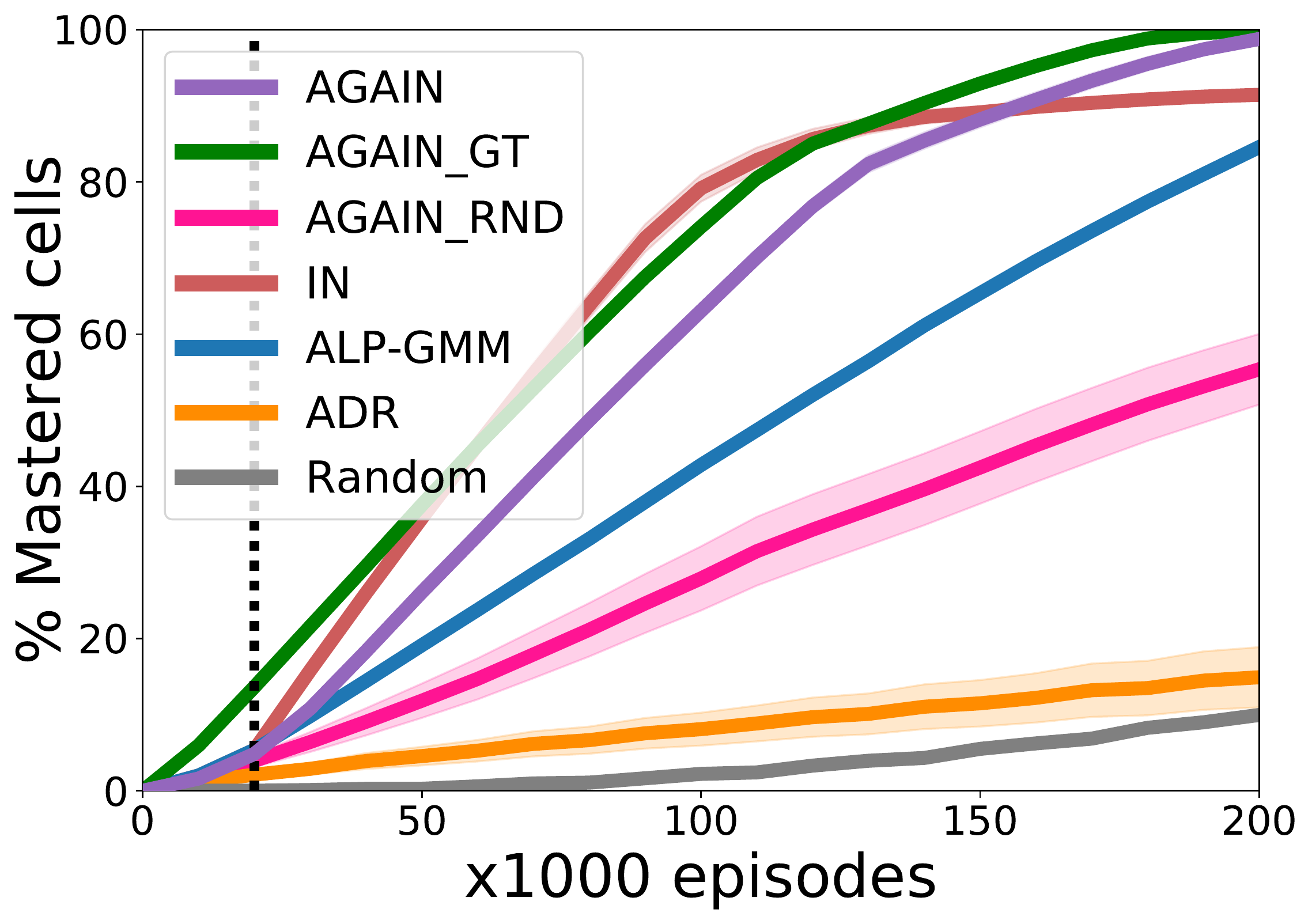}}
\subfloat{\includegraphics[width=0.45\textwidth]{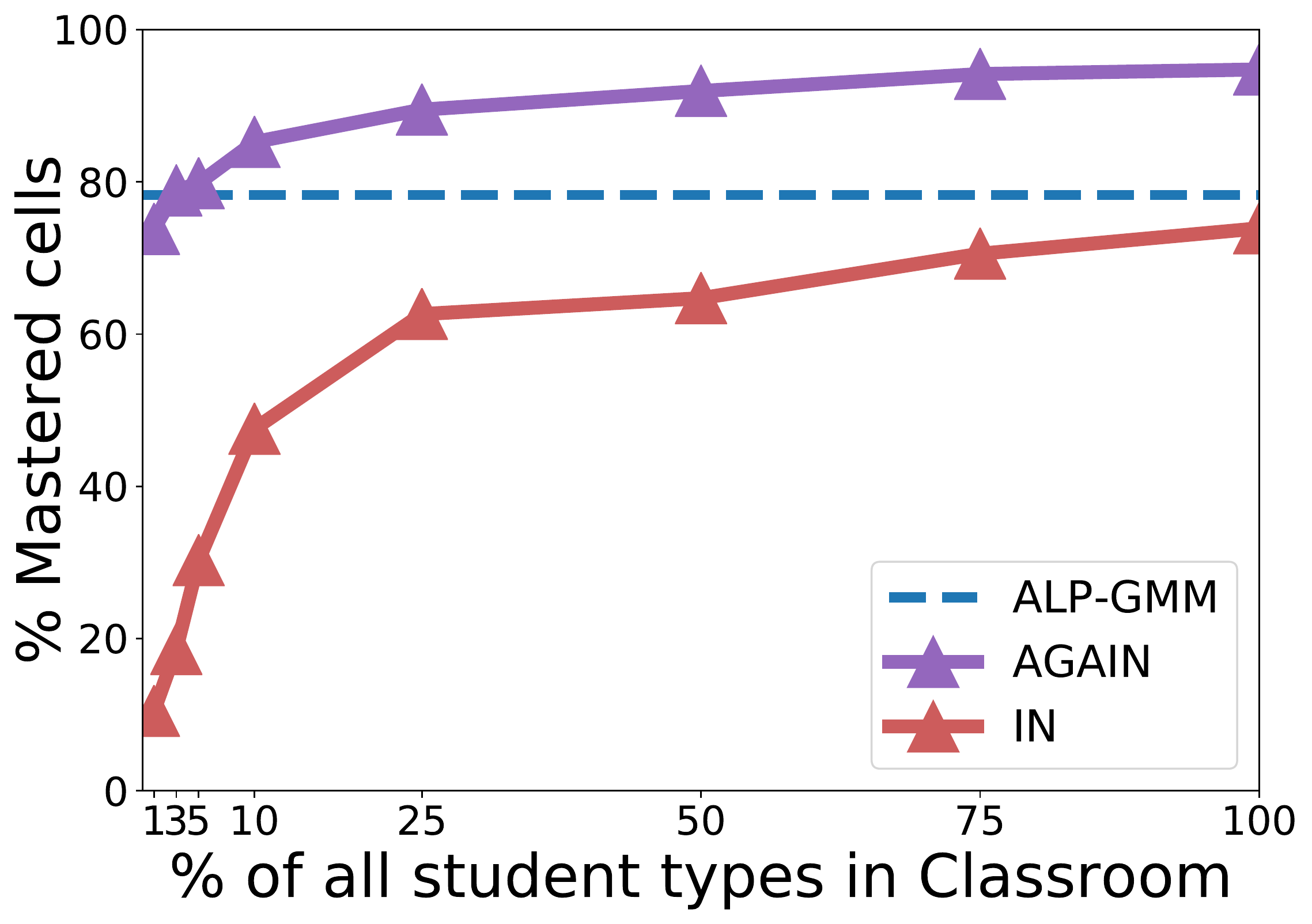}}
\caption{\footnotesize{\textbf{left:} By leveraging meta-learned curriculum priors w.r.t to its students, AGAIN outperforms regular ACL approaches. Avg. perfs. with \textit{sem} (standard error of the mean) plotted, 48 seeds. The vertical dashed black line indicates when pre-training ends for Meta-ACL conditions. \textbf{right:} Impact of classroom size and sparsity on Meta-ACL performances. Post-training ($200$k ep.) avg perfs. plotted, 96 seeds.}}
\label{vizu-toy-env-classroom}
\end{figure*}

\paragraph{Results} Instead of performing a pre-test to construct the KC vector of a student, we directly compute it by concatenating $|C|$ for all cells, giving a $400$-dimensional KC vector. This vector is computed after $20$k training episodes out of $200$k. To study AGAIN, we first populate our training trajectory history $\mathcal{H}$ by training with ALP-GMM an initial classroom of 128 students drawn randomly from 4 fixed possible student types (i.e. 4 possible initially unlocked cell positions), and then test it on a new fixed set of $48$ random students.

\textit{Comparative analysis - } Figure \ref{vizu-toy-env-classroom} (left) showcases performance across training for our considered Meta-ACL conditions and ACL baselines. Both AGAIN and IN significantly outperform ALP-GMM ($p<.001$ for both, using Welch's t-test at $200$k episodes). The initial performance advantage of IN w.r.t AGAIN is due to the greedy nature of IN, which only exploits the expert curriculum while AGAIN complements it with ALP-GMM for exploration. By the end of training, AGAIN outperforms IN ($p<.001$) thanks to its ability to emancipate from the curriculum priors it initially leverages. The regular KC-based curriculum priors selection used in AGAIN outperformed the random selection used in AGAIN\_RND ($p<.001$ at $200$k episodes), while being not significantly inferior to the Ground Truth variant AGAIN\_GT ($p=0.16$). Because we assume no expert knowledge over the set of students to train, i.e. their respective initial learning subspace is unknown, ADR -- which relies on being given an initial easy task -- fails to train most students when given randomly selected starting subspace (among the $4$ possible ones). By contrast, this showcases the ability of AGAIN to autonomously and efficiently infer such expert knowledge.

\textit{Varying classroom size experiment - } An important property that must be met by a meta-learning procedure is to have a monotonic increase of performance as the database of information being leveraged increases. Another important expected aspect of Meta-ACL is whether the approach is able to generalize to students that were never seen before. To assess whether these properties hold on AGAIN, we consider the full student distribution of the toy environment, i.e. $400$ possible student types. We populate a new history $\mathcal{H}$ by training (with ALP-GMM) a $400$-students classroom (one per student type). We then analyse the end performance of AGAIN and IN on a fixed test set of 96 random students when given increasingly smaller subsets of $\mathcal{H}$. The smaller the subset, the harder it becomes to generalize over new students. Results, shown in fig. \ref{vizu-toy-env-classroom} (right), demonstrate that both AGAIN and IN do have monotonic performance increasements as the classroom grows. With as little as $10$\% of possible students in the classroom, AGAIN statistically significantly ($p<.001$) outperforms ALP-GMM on the new student set, i.e. it generalizes to never seen before students.

 \subsection{Meta-ACL for DRL students in the Parkour environment}
 \label{sec:exp:walker-climber}
 
 \begin{wrapfigure}[14]{r}{6.3cm}
\vspace{-0.4cm}
\includegraphics[width=5.7cm]{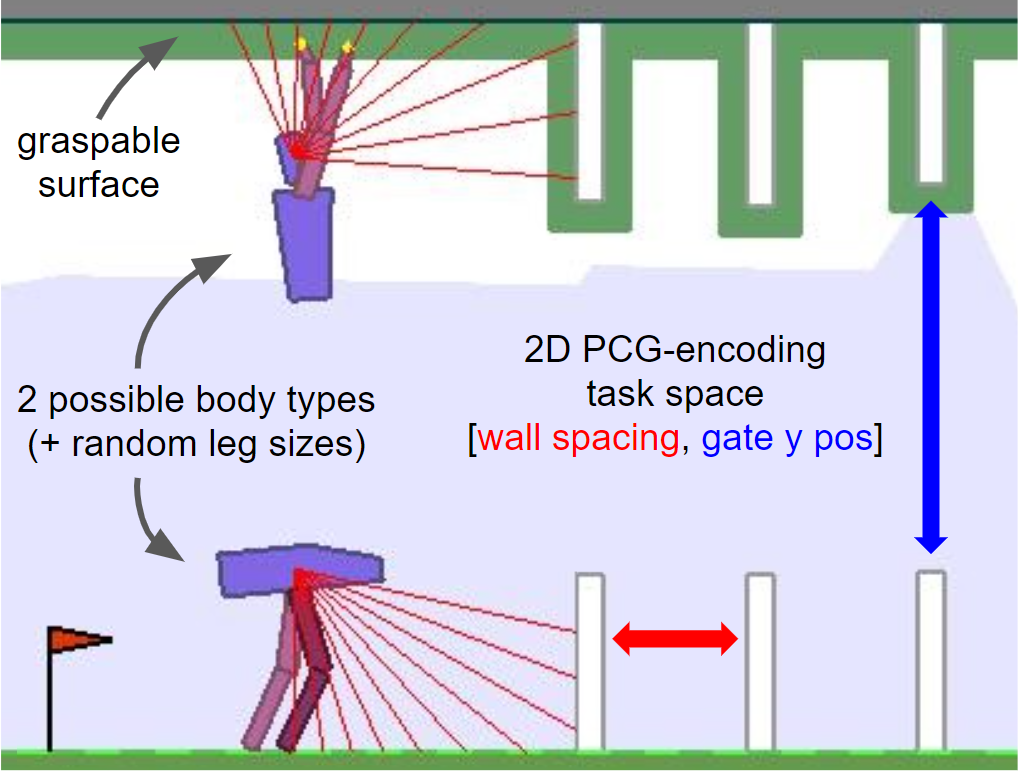}
\caption{\footnotesize{Our proposed parametric Parkour env. to study Meta-ACL with DRL students.}}
\label{parkour-env}
\end{wrapfigure}
 
To study Meta-ACL with DRL students, we present a Box2D Parkour environment with a $2$D parametric PCG that encodes a large space of tasks (see fig. \ref{parkour-env}). The first parameter controls the spacing between walls that are positioned along the track, while the second parameter sets the y-position of a gate that is added to each wall. Positive rewards are collected by going forward. To simulate a multi-modal distribution of students well suited to study Meta-ACL, we randomize the student's morphology for each new training (i.e. each seed): It can be embodied in either a bipedal walker, which will be prone to learn tasks with near-ground gate positions, or a two-armed climber, for which tasks with near-roof gate positions are easiest. We also randomize the student's limb sizes which can vary from the length visible in fig. \ref{parkour-env} to 50\% shorter.

\paragraph{Results} In the following experiments our Meta-ACL variants leverage a history $\mathcal{H}$ built from a classroom of $128$ randomly drawn Soft-Actor-Critic \citep{sac} students (i.e. varying embodiments and initial policy weights) trained with ALP-GMM. We then compare ACL and Meta-ACL variants on a fixed set of 64 new students and report the mean percentage of mastered environments (i.e. $r>230$) from 2 fixed expert test sets (one per embodiment type) across training. The KC vector is built using a uniform pre-test set of $m=225$ tasks, performed after $2$ millions agent steps out of $10$. See appendix \ref{ann:parkour} for additional experimental details.

\begin{figure*}[htb!]
\centering
\subfloat{\includegraphics[width=0.49\textwidth]{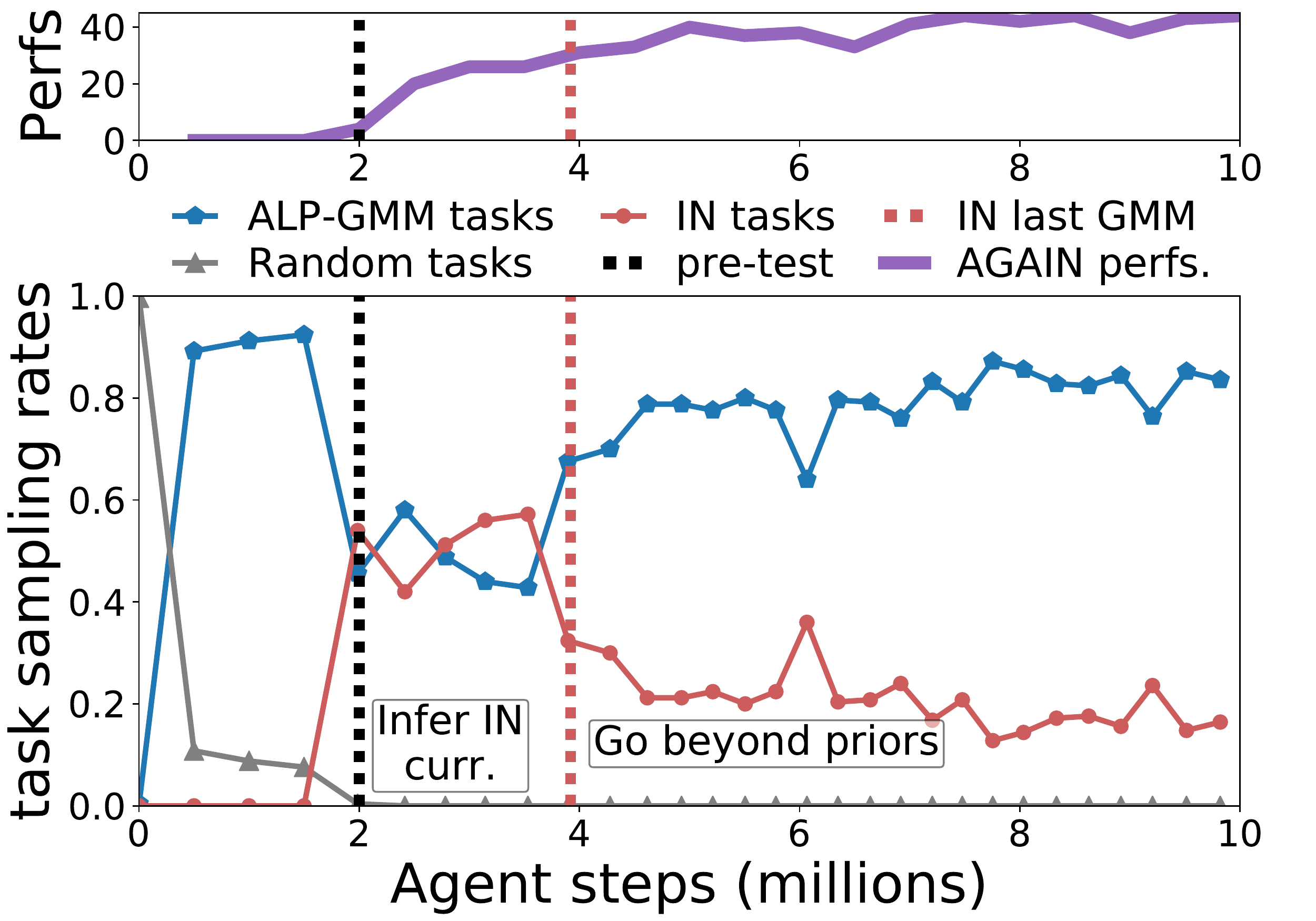}}
\subfloat{\includegraphics[width=0.48\textwidth]{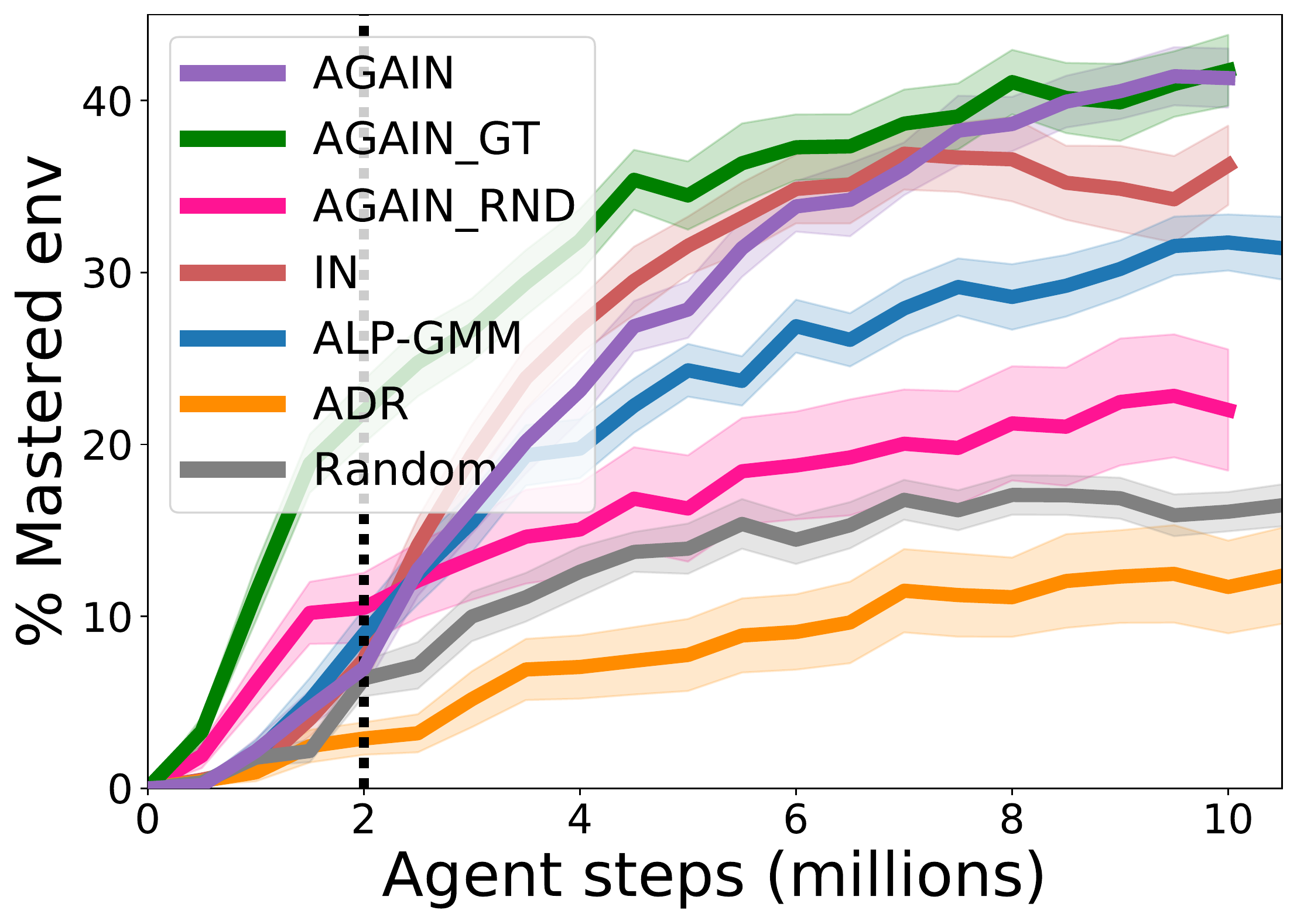}}
\caption{\footnotesize{\textbf{left:} Example of evolution of task sampling when using AGAIN in the Parkour env. (1 seed). \textbf{right:} Average performances of AGAIN with variants and baselines in the Parkour env.. 64 seeds, sem plotted. The vertical dashed black line indicates when pre-training ends for Meta-ACL conditions.}}
\label{wc-env-vizu}
\end{figure*}

\textit{Qualitative view - } Figure \ref{wc-env-vizu} (left) showcases the evolution of task sampling when using AGAIN to train a new student. Three distinct phases emerge: 1) A pre-training exploratory phase used to gather information about the student’s capabilities, 2) After building the KC vector and inferring the most appropriate curriculum priors from $\mathcal{H}$, AGAIN paces through the resulting IN curriculum while mixing it to ALP-GMM, and 3) AGAIN emancipates from IN after completing it.

\textit{Comparative analysis - } As shown in figure \ref{wc-env-vizu} (right), through its use of curriculum priors, AGAIN outperforms ALP-GMM on Parkour, mastering an average of $41\%$ of the test set at $10$M steps, compared to $31\%$ for ALP-GMM ($p<.001$) after $10.5$M steps ($0.5$M training steps added to account for AGAIN additional pre-test time). AGAIN performs better than its AGAIN\_RND random prior selection variant, and is not statistically different ($p=0.8$) from ground truth sampling (AGAIN\_GT), although only by the end of training. While AGAIN and IN initially have comparable performances, after $7$ Millions training steps, -- a point at which most students trained with IN or AGAIN reached the last IN GMM --, AGAIN outperforms IN by the end of training ($p<0.02$). This showcases the advantage of emancipating from the expert curriculum once completed. As in the toy environment experiments, when given randomly selected starting subspaces (since we assume no expert knowledge), ADR fails to train most students.

 \subsection{Applying Meta-ACL to a single student: Trying AGAIN instead of trying longer}
 \label{exp:trying-again}

Given a single DRL student to train (i.e. no history $\mathcal{H}$), and if we do not assume access to expert knowledge, current ACL approaches leverage task-exploration (as in ALP-GMM). We hypothesize that these additional tasks presented to the DRL learner have a cluttering effect on the gathered training data, which adds noise in its already brittle gradient-based optimization and leads to sub-optimal performances. We propose to address this problem by modifying AGAIN to fit this no-history setup and by allowing to restart the student along training. More precisely, instead of pre-testing the student to find appropriate curriculum priors in $\mathcal{H}$, we split the training of the target student into a two stage approach where 1) the DRL student is first trained with ALP-GMM (with high-exploration), and then 2) we extract curriculum priors from the training trajectory of the first run and use them to re-train the same agent \textit{from scratch}.
 
 \begin{wrapfigure}[18]{r}{6.5cm}
 \vspace{-0.2cm}
\includegraphics[width=6.5cm]{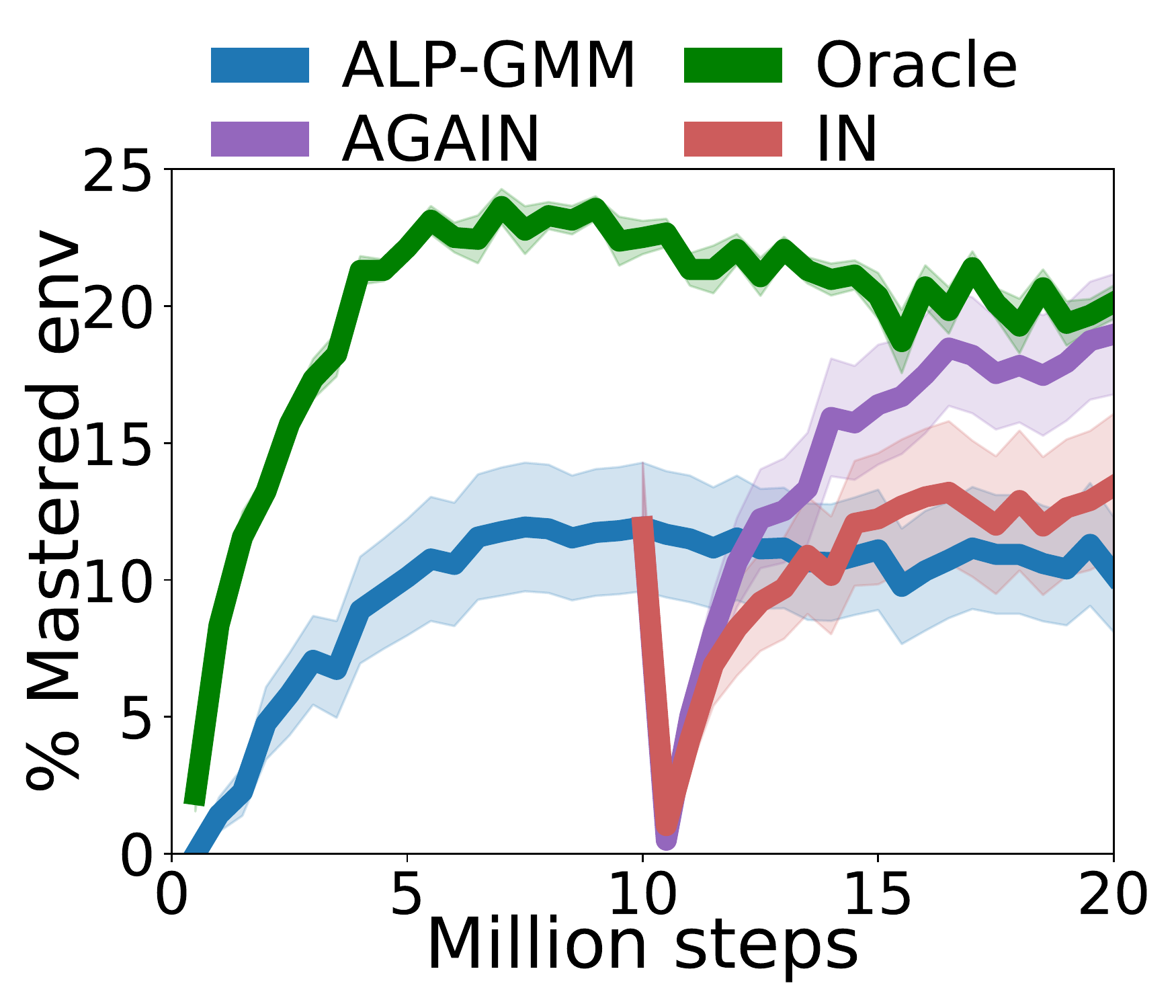}
\caption{\footnotesize{Given a single DRL student to train, AGAIN outperforms ALP-GMM in a parametric BipedalWalker environment. sem plotted, 32 seeds.}}\label{trying-again-compact}
\end{wrapfigure} 
\paragraph{Results.} We test our modified AGAIN along with variants and baselines on a parametric version of BipedalWalker proposed in \citep{portelas2019},  which generates walking tracks paved with stumps whose height and spacing are defined by a PCG-encoding $2$-D parameter vector. As in their work, we test our approaches with both the default walker and a modified short-legged walker, which constitutes an even more challenging scenario (as the task space is unchanged). Performance is measured by tracking the percentage of mastered tasks from a fixed test set. See App. \ref{ann:tryingagain} for a complete analysis. 

Figure \ref{trying-again-compact} showcases our proposed approach on the short walker setup (with a SAC student \citep{sac}). On this short walker scenario, mixing ALP-GMM with IN is essential: while IN end performances are not statistically significantly superior to ALP-GMM, AGAIN clearly outperforms ALP-GMM $(p<0.01)$, reaching a mean end performance of $19.0$. The difference in end-performance between AGAIN and Oracle, our hand-made curriculum using privileged information who obtained $20.1$, is not significant ($p=0.6$).

\section{Conclusion and Discussion}
In this work we attempted to motivate and formalize the study of Classroom Teaching problems, in which a set of diverse students have to be trained optimally, and we proposed to attain this goal through the use of Meta-ACL algorithms. We then presented AGAIN, a first Meta-ACL baseline, and demonstrated its advantages over classical ACL and variants for CT problems in both a toy environment and in a new parametric Parkour environment with DRL learners. We also showed how AGAIN can bring performance gains over ACL in classical single student ACL scenarios.

\paragraph{Future work}In future work, AGAIN could be improved by using adaptive approaches to build compact pre-test sets, e.g. using decision tree based test pruning methods, or by combining curriculum priors from multiple previously trained learners. While AGAIN is built on top of an existing ACL algorithm, developing an end-to-end Meta-ACL algorithm that generates curricula using a DRL teacher policy trained across multiple students is also a promising line of work to follow. 
Additionally, this work opens-up exciting new perspectives in transferring Meta-ACL methods to educational data-mining, e.g. in MOOC scenarios, given a previously trained pilot classroom, one could use Meta-ACL to infer adaptive curricula for new students.

\paragraph{Potential negative societal impact} Meta Automatic Curriculum Learning exploits previous curriculum data of DRL students to better train new ones. As such, if previously trained students acquired unwanted biases, they could potentially be transferred and further amplified from the Meta-ACL training of new DRL students. This can have serious negative societal impacts if considering applications in socially impactful domains (e.g. deciding whether to give an insurance to someone). Further work will need to study how one
can safely learn curriculum priors that are not harmful in sensitive application areas.

\section*{Acknowledgments}
This work was supported by Microsoft Research through its PhD Scholarship Programme. All presented experiments were carried out using 1) the computing facilities MCIA (Mésocentre de Calcul Intensif Aquitain) of the Université de Bordeaux and of the Université de Pau et des Pays de l'Adour, and 2) the HPC resources of IDRIS under the allocation 2020-[A0091011996] made by GENCI.

\clearpage
\bibliography{biblio}  

\begin{thebibliography}{48}
\providecommand{\natexlab}[1]{#1}
\providecommand{\url}[1]{\texttt{#1}}
\expandafter\ifx\csname urlstyle\endcsname\relax
  \providecommand{\doi}[1]{doi: #1}\else
  \providecommand{\doi}{doi: \begingroup \urlstyle{rm}\Url}\fi

\bibitem[Selfridge et~al.(1985)Selfridge, Sutton, and Barto]{selfridge}
Oliver~G. Selfridge, Richard~S. Sutton, and Andrew~G. Barto.
\newblock Training and tracking in robotics.
\newblock In \emph{{IJCAI}}, 1985.

\bibitem[Schmidhuber(1991)]{Schmid}
Jürgen Schmidhuber.
\newblock Curious model-building control systems.
\newblock In \emph{IJCNN}. IEEE, 1991.

\bibitem[Oudeyer et~al.(2007)Oudeyer, Kaplan, and Hafner]{oudeyer2007intrinsic}
Pierre-Yves Oudeyer, Frdric Kaplan, and Verena~V Hafner.
\newblock Intrinsic motivation systems for autonomous mental development.
\newblock \emph{IEEE trans. on evolutionary comp.}, 2007.

\bibitem[Elman(1993)]{elman}
Jeffrey~L. Elman.
\newblock Learning and development in neural networks: the importance of
  starting small.
\newblock \emph{Cognition}, 48\penalty0 (1):\penalty0 71 -- 99, 1993.
\newblock ISSN 0010-0277.

\bibitem[Bengio et~al.(2009)Bengio, Louradour, Collobert, and Weston]{bengiocl}
Yoshua Bengio, J{\'{e}}r{\^{o}}me Louradour, Ronan Collobert, and Jason Weston.
\newblock Curriculum learning.
\newblock In \emph{{ICML}}, 2009.

\bibitem[Portelas et~al.(2020)Portelas, Colas, Weng, Hofmann, and
  Oudeyer]{portelas2020-acl-drl}
Rémy Portelas, Cédric Colas, Lilian Weng, Katja Hofmann, and Pierre-Yves
  Oudeyer.
\newblock Automatic curriculum learning for deep rl: A short survey.
\newblock \emph{{IJCAI}}, 2020.

\bibitem[Pathak et~al.(2017)Pathak, Agrawal, Efros, and Darrell]{icm}
Deepak Pathak, Pulkit Agrawal, Alexei~A Efros, and Trevor Darrell.
\newblock Curiosity-driven exploration by self-supervised prediction.
\newblock In \emph{CVPR}, 2017.

\bibitem[Burda et~al.(2019)Burda, Edwards, Storkey, and Klimov]{rnd}
Yuri Burda, Harrison Edwards, Amos~J. Storkey, and Oleg Klimov.
\newblock Exploration by random network distillation.
\newblock \emph{ICLR}, 2019.

\bibitem[Salimans and Chen(2018)]{montezuma-single-demo}
Tim Salimans and Richard Chen.
\newblock Learning montezuma's revenge from a single demonstration.
\newblock \emph{NeurIPS}, 2018.

\bibitem[Andrychowicz et~al.(2017)Andrychowicz, Wolski, Ray, Schneider, Fong,
  Welinder, McGrew, Tobin, Abbeel, and Zaremba]{her}
Marcin Andrychowicz, Filip Wolski, Alex Ray, Jonas Schneider, Rachel Fong,
  Peter Welinder, Bob McGrew, Josh Tobin, OpenAI~Pieter Abbeel, and Wojciech
  Zaremba.
\newblock Hindsight experience replay.
\newblock In \emph{NeurIPS}, 2017.

\bibitem[Colas et~al.(2019)Colas, Oudeyer, Sigaud, Fournier, and
  Chetouani]{curious}
C{\'e}dric Colas, Pierre-Yves Oudeyer, Olivier Sigaud, Pierre Fournier, and
  Mohamed Chetouani.
\newblock Curious: Intrinsically motivated modular multi-goal reinforcement
  learning.
\newblock In \emph{ICML}, 2019.

\bibitem[Cideron et~al.(2019)Cideron, Seurin, Strub, and
  Pietquin]{cideron2019self}
Geoffrey Cideron, Mathieu Seurin, Florian Strub, and Olivier Pietquin.
\newblock Self-educated language agent with hindsight experience replay for
  instruction following.
\newblock \emph{ViGIL, NeurIPS Workshop}, 2019.

\bibitem[Fournier et~al.(2018)Fournier, Sigaud, Chetouani, and
  Oudeyer]{fournier-accuracy-acl}
Pierre Fournier, Olivier Sigaud, Mohamed Chetouani, and Pierre{-}Yves Oudeyer.
\newblock Accuracy-based curriculum learning in deep reinforcement learning.
\newblock \emph{arXiv}, 2018.

\bibitem[Matiisen et~al.(2017)Matiisen, Oliver, Cohen, and Schulman]{tscl}
Tambet Matiisen, Avital Oliver, Taco Cohen, and John Schulman.
\newblock Teacher-student curriculum learning.
\newblock \emph{IEEE TNNLS}, 2017.

\bibitem[Portelas et~al.(2019)Portelas, Colas, Hofmann, and
  Oudeyer]{portelas2019}
Rémy Portelas, Cédric Colas, Katja Hofmann, and Pierre-Yves Oudeyer.
\newblock Teacher algorithms for curriculum learning of deep rl in continuously
  parameterized environments.
\newblock \emph{CoRL}, 2019.

\bibitem[Mehta et~al.(2019)Mehta, Diaz, Golemo, Pal, and Paull]{ADRmila}
Bhairav Mehta, Manfred Diaz, Florian Golemo, Christopher~J. Pal, and Liam
  Paull.
\newblock Active domain randomization.
\newblock \emph{CoRL}, 2019.

\bibitem[Florensa et~al.(2018)Florensa, Held, Geng, and Abbeel]{goalgan}
Carlos Florensa, David Held, Xinyang Geng, and Pieter Abbeel.
\newblock Automatic goal generation for reinforcement learning agents.
\newblock In \emph{ICML}, 2018.

\bibitem[Klink et~al.(2020)Klink, D'Eramo, Peters, and Pajarinen]{selfpaceddrl}
Pascal Klink, Carlo D'Eramo, Jan Peters, and Joni Pajarinen.
\newblock Self-paced deep reinforcement learning.
\newblock \emph{arXiv}, 2020.

\bibitem[Risi and Togelius(2019)]{risiPCG}
Sebastian Risi and Julian Togelius.
\newblock Procedural content generation: From automatically generating game
  levels to increasing generality in machine learning.
\newblock \emph{arXiv}, 2019.

\bibitem[Justesen et~al.(2018)Justesen, Torrado, Bontrager, Khalifa, Togelius,
  and Risi]{illuminating}
Niels Justesen, Ruben~Rodriguez Torrado, Philip Bontrager, Ahmed Khalifa,
  Julian Togelius, and Sebastian Risi.
\newblock Illuminating generalization in deep reinforcement learning through
  procedural level generation.
\newblock \emph{NeurIPS Deep RL Workshop}, 2018.

\bibitem[Cobbe et~al.(2019)Cobbe, Klimov, Hesse, Kim, and
  Schulman]{quantif-coinrun}
Karl Cobbe, Oleg Klimov, Christopher Hesse, Taehoon Kim, and John Schulman.
\newblock Quantifying generalization in reinforcement learning.
\newblock \emph{ICML}, abs/1812.02341, 2019.

\bibitem[Mnih et~al.(2015)Mnih, Kavukcuoglu, Silver, Rusu, Veness, Bellemare,
  Graves, Riedmiller, Fidjeland, Ostrovski, et~al.]{dqn}
Volodymyr Mnih, Koray Kavukcuoglu, David Silver, Andrei~A Rusu, Joel Veness,
  Marc~G Bellemare, Alex Graves, Martin Riedmiller, Andreas~K Fidjeland, Georg
  Ostrovski, et~al.
\newblock Human-level control through deep reinforcement learning.
\newblock \emph{Nature}, 518\penalty0 (7540):\penalty0 529, 2015.

\bibitem[Schulman et~al.(2015)Schulman, Levine, Abbeel, Jordan, and
  Moritz]{trpo}
John Schulman, Sergey Levine, Pieter Abbeel, Michael~I. Jordan, and Philipp
  Moritz.
\newblock Trust region policy optimization.
\newblock In \emph{ICML}, 2015.

\bibitem[Lillicrap et~al.(2016)Lillicrap, Hunt, Pritzel, Heess, Erez, Tassa,
  Silver, and Wierstra]{ddpg}
Timothy~P. Lillicrap, Jonathan~J. Hunt, Alexander Pritzel, Nicolas Heess, Tom
  Erez, Yuval Tassa, David Silver, and Daan Wierstra.
\newblock Continuous control with deep reinforcement learning.
\newblock In \emph{ICLR}, 2016.

\bibitem[Forestier et~al.(2017)Forestier, Portelas, Mollard, and
  Oudeyer]{imgep}
S{\'{e}}bastien Forestier, R{\'{e}}my Portelas, Yoan Mollard, and Pierre{-}Yves
  Oudeyer.
\newblock Intrinsically motivated goal exploration processes with automatic
  curriculum learning.
\newblock \emph{arXiv}, 2017.

\bibitem[Wang et~al.(2019)Wang, Lehman, Clune, and Stanley]{poet}
Rui Wang, Joel Lehman, Jeff Clune, and Kenneth~O. Stanley.
\newblock Paired open-ended trailblazer {(POET):} endlessly generating
  increasingly complex and diverse learning environments and their solutions.
\newblock \emph{arXiv}, 2019.

\bibitem[Cl{\'e}ment et~al.(2015)Cl{\'e}ment, Roy, Oudeyer, and Lopes]{zpdes}
Benjamin Cl{\'e}ment, Didier Roy, Pierre-Yves Oudeyer, and Manuel Lopes.
\newblock {Multi-Armed Bandits for Intelligent Tutoring Systems}.
\newblock \emph{{Journal of Educational Data Mining (JEDM)}}, 7\penalty0
  (2):\penalty0 20--48, June 2015.

\bibitem[Koedinger et~al.(2013)Koedinger, Brunskill, de~Baker, McLaughlin, and
  Stamper]{Koedinger13}
Kenneth~R. Koedinger, Emma Brunskill, Ryan Shaun~Joazeiro de~Baker,
  Elizabeth~A. McLaughlin, and John~C. Stamper.
\newblock New potentials for data-driven intelligent tutoring system
  development and optimization.
\newblock \emph{AI Magazine}, 34\penalty0 (3):\penalty0 27--41, 2013.

\bibitem[Mysore et~al.(2019)Mysore, Platt, and Saenko]{tscllike}
S.~Mysore, R.~Platt, and K.~Saenko.
\newblock Reward-guided curriculum for robust reinforcement learning.
\newblock \emph{Workshop on Multi-task and Lifelong Reinforcement Learning at
  ICML}, 2019.

\bibitem[Eysenbach et~al.(2018)Eysenbach, Gupta, Ibarz, and Levine]{diayn}
Benjamin Eysenbach, Abhishek Gupta, Julian Ibarz, and Sergey Levine.
\newblock Diversity is all you need: Learning skills without a reward function.
\newblock \emph{arXiv}, 2018.

\bibitem[Jabri et~al.(2019)Jabri, Hsu, Gupta, Eysenbach, Levine, and
  Finn]{metarl-carml}
Allan Jabri, Kyle Hsu, Abhishek Gupta, Ben Eysenbach, Sergey Levine, and
  Chelsea Finn.
\newblock Unsupervised curricula for visual meta-reinforcement learning.
\newblock In \emph{NeurIPS}. 2019.

\bibitem[Bellemare et~al.(2016)Bellemare, Srinivasan, Ostrovski, Schaul,
  Saxton, and Munos]{countbased}
Marc Bellemare, Sriram Srinivasan, Georg Ostrovski, Tom Schaul, David Saxton,
  and Remi Munos.
\newblock Unifying count-based exploration and intrinsic motivation.
\newblock In \emph{NeurIPS}, 2016.

\bibitem[Racanière et~al.(2020)Racanière, Lampinen, Santoro, Reichert,
  Firoiu, and Lillicrap]{settersolver}
Sébastien Racanière, Andrew Lampinen, Adam Santoro, David Reichert, Vlad
  Firoiu, and Timothy Lillicrap.
\newblock Automated curricula through setter-solver interactions.
\newblock \emph{ICLR}, 2020.

\bibitem[OpenAI et~al.(2019)OpenAI, Akkaya, Andrychowicz, Chociej, Litwin,
  McGrew, Petron, Paino, Plappert, Powell, Ribas, Schneider, Tezak, Tworek,
  Welinder, Weng, Yuan, Zaremba, and Zhang]{OpenAI2019SolvingRC}
OpenAI, Ilge Akkaya, Marcin Andrychowicz, Maciek Chociej, Mateusz Litwin, Bob
  McGrew, Arthur Petron, Alex Paino, Matthias Plappert, Glenn Powell, Raphael
  Ribas, Jonas Schneider, Nikolas Tezak, Jadwiga Tworek, Peter Welinder, Lilian
  Weng, Qi-Ming Yuan, Wojciech Zaremba, and Lefei Zhang.
\newblock Solving rubik's cube with a robot hand.
\newblock \emph{ArXiv}, 2019.

\bibitem[Florensa et~al.(2017)Florensa, Held, Wulfmeier, and
  Abbeel]{reverse-cur}
Carlos Florensa, David Held, Markus Wulfmeier, and Pieter Abbeel.
\newblock Reverse curriculum generation for reinforcement learning.
\newblock \emph{CoRL}, 2017.

\bibitem[Teh et~al.(2017)Teh, Bapst, Czarnecki, Quan, Kirkpatrick, Hadsell,
  Heess, and Pascanu]{distralTeh2017}
Yee~Whye Teh, Victor Bapst, Wojciech Czarnecki, John Quan, James Kirkpatrick,
  Raia Hadsell, Nicolas Manfred~Otto Heess, and Razvan Pascanu.
\newblock Distral: Robust multitask reinforcement learning.
\newblock In \emph{NIPS}, 2017.

\bibitem[Czarnecki et~al.(2019)Czarnecki, Pascanu, Osindero, Jayakumar,
  Swirszcz, and Jaderberg]{pol-dil-review}
Wojciech~Marian Czarnecki, Razvan Pascanu, Simon Osindero, Siddhant~M.
  Jayakumar, Grzegorz Swirszcz, and Max Jaderberg.
\newblock Distilling policy distillation.
\newblock \emph{AISTATS}, 2019.

\bibitem[Hacohen and Weinshall(2019)]{hacohen19a-scoring-pacing}
Guy Hacohen and Daphna Weinshall.
\newblock On the power of curriculum learning in training deep networks.
\newblock In Kamalika Chaudhuri and Ruslan Salakhutdinov, editors, \emph{ICML},
  2019.

\bibitem[Furlanello et~al.(2018)Furlanello, Lipton, Tschannen, Itti, and
  Anandkumar]{ban}
Tommaso Furlanello, Zachary~Chase Lipton, Michael Tschannen, Laurent Itti, and
  Anima Anandkumar.
\newblock Born-again neural networks.
\newblock In \emph{ICML}, pages 1602--1611, 2018.

\bibitem[Yim et~al.(2017)Yim, Joo, Bae, and Kim]{banlike}
Junho Yim, Donggyu Joo, Jihoon Bae, and Junmo Kim.
\newblock A gift from knowledge distillation: Fast optimization, network
  minimization and transfer learning.
\newblock \emph{CVPR}, pages 7130--7138, 2017.

\bibitem[Turchetta et~al.(2020)Turchetta, Kolobov, Shah, Krause, and
  Agarwal]{safe-rl-curr-induction}
Matteo Turchetta, Andrey Kolobov, Shital Shah, Andreas Krause, and Alekh
  Agarwal.
\newblock Safe reinforcement learning via curriculum induction.
\newblock \emph{ArXiv}, 2020.

\bibitem[Vanschoren(2018)]{surveymetarl}
Joaquin Vanschoren.
\newblock Meta-learning: {A} survey.
\newblock \emph{arXiv}, 2018.

\bibitem[Wang et~al.(2016)Wang, Kurth{-}Nelson, Tirumala, Soyer, Leibo, Munos,
  Blundell, Kumaran, and Botvinick]{wang2016}
Jane~X. Wang, Zeb Kurth{-}Nelson, Dhruva Tirumala, Hubert Soyer, Joel~Z. Leibo,
  R{\'{e}}mi Munos, Charles Blundell, Dharshan Kumaran, and Matthew Botvinick.
\newblock Learning to reinforcement learn.
\newblock \emph{arXiv}, 2016.

\bibitem[Auer et~al.(2002)Auer, Cesa-Bianchi, Freund, and
  Schapire]{auer2002nonstochastic}
Peter Auer, Nicolo Cesa-Bianchi, Yoav Freund, and Robert~E Schapire.
\newblock The nonstochastic multiarmed bandit problem.
\newblock \emph{SIAM journal on computing}, 32\penalty0 (1):\penalty0 48--77,
  2002.

\bibitem[Vie et~al.(2018)Vie, Popineau, Bruillard, and
  Bourda]{vie-mooc-test-assembly}
Jill-J{\^e}nn Vie, Fabrice Popineau, {\'E}ric Bruillard, and Yolaine Bourda.
\newblock {Automated Test Assembly for Handling Learner Cold-Start in
  Large-Scale Assessments}.
\newblock \emph{{IJAIED}}, 2018.

\bibitem[Vie(2016)]{vie-thesis}
Jill-J{\^e}nn Vie.
\newblock \emph{{Cognitive diagnostic computerized adaptive testing models for
  large-scale learning}}.
\newblock Thesis, {Universit{\'e} Paris Saclay (COmUE)}, 2016.

\bibitem[Haarnoja et~al.(2018)Haarnoja, Zhou, Abbeel, and Levine]{sac}
Tuomas Haarnoja, Aurick Zhou, Pieter Abbeel, and Sergey Levine.
\newblock Soft actor-critic: Off-policy maximum entropy deep reinforcement
  learning with a stochastic actor.
\newblock \emph{ICML}, 2018.

\bibitem[Bozdogan(1987)]{aic}
Hamparsum Bozdogan.
\newblock Model selection and akaike's information criterion (aic): The general
  theory and its analytical extensions.
\newblock \emph{Psychometrika}, 52\penalty0 (3):\penalty0 345--370, Sep 1987.

\end{thebibliography}
\bibliographystyle{unsrtnat}

\clearpage

\appendix

\renewcommand{\algorithmiccomment}[1]{#1}

\section{ALP-GMM}
\label{app-alp-gmm}
\begin{figure*}[ht!]
\centering
\includegraphics[width=0.80\textwidth]{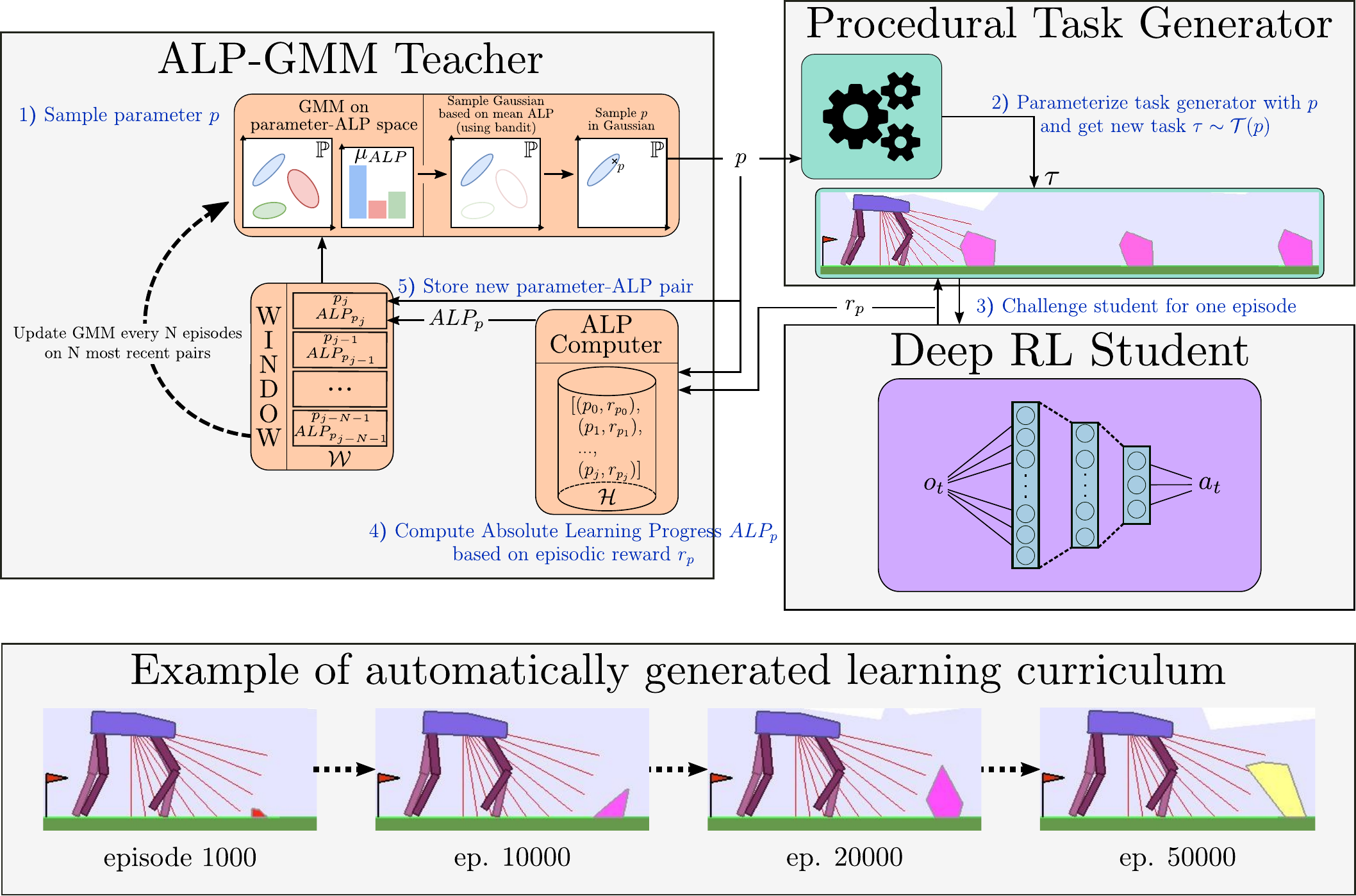}
\caption{\footnotesize{Schematic view of an ALP-GMM teacher's workflow from \cite{portelas2019}}}
\label{ALP-GMM-pipeline}
\end{figure*}

ALP-GMM \cite{portelas2019} (MIT license) relies on an empirical per-task computation of Absolute Learning Progress (ALP), allowing to fit a GMM on a concatenated space composed of tasks' parameters and respective ALP. Given a task $a_{new} \in \mathcal{A}$ whose parameter is $p_{new} \in \mathcal{P}$ and on which the student's policy collected the episodic reward $r_{new} \in \mathbb{R}$, Its ALP is computed using the closest previous tasks $a_{old}$ (Euclidean distance) with associated episodic reward $r_{old}$:
\begin{equation}
    \label{eq:2}
    alp_{new} = |r_{new} - r_{old}|
\end{equation}
All previously encountered task's parameters and their associated ALP, parameter-ALP for short, recorded in a history database $H$, are used for this computation. Contrastingly, the fitting of the GMM is performed every $N$ episodes on a window $\mathcal{W}$ containing the $N$ most recent parameter-ALP. The resulting mean ALP dimension of each Gaussian of the GMM is used for proportional sampling. To adapt the number of components of the GMM online, a batch of GMMs having from 2 to $k_{max}$ components is fitted on $\mathcal{W}$, and the best one, according to Akaike's Information Criterion \cite{aic}, is kept as the new GMM. In all of our experiments we use the same hyperparameters as in \cite{portelas2019} ($N=250$, $k_{max}=10$), except for the percentage of random task sampling $\rho_{rnd}$ which we set to $10\%$ (we found it to perform better than $20\%$) when running ALP-GMM. See algorithm \ref{algo:ALP-GMM} for pseudo-code and figure \ref{ALP-GMM-pipeline} for a schematic pipeline. Note that in the main body of this paper we refer to ALP as LP for simplicity (ie. $LP_{ti}$ in $\mathcal{C}$ from eq. \ref{eq:craw} is equivalent to the mean ALP of Gaussians in ALP-GMM).


\begin{algorithm*}[htb!]
	\caption{~ Absolute Learning Progress Gaussian Mixture Model (ALP-GMM)}
	\label{algo:ALP-GMM}
	\begin{algorithmic}
	
	\REQUIRE Student policy $s_\theta$, parametric procedural environment generator $E$, bounded parameter space $\mathcal{P}$, probability of random sampling $\rho_{rnd}$, fitting rate $N$, max number of Gaussians $k_{max}$
	\vspace{0.2cm}
	\STATE Initialize $s_\theta$
	\STATE Initialize parameter-ALP First-in-First-Out window $\mathcal{W}$, set max size to $N$
	\STATE Initialize parameter-reward history database $H$
	\LOOP[{$N$ times ~~~~{\color{gray}~~~\#  Bootstrap phase}}]
	    \STATE Sample random $p \in \mathcal{P}$, send $E(a \sim \mathcal{A}(p))$ to $s_\theta$, observe episodic reward $r_p$
	    \STATE Compute ALP of $p$ based on $r_p$ and $H$ (see equation \ref{eq:2})
	    \STATE Store $(p,r_p)$ pair in $H$, store $(p,ALP_p)$ pair in $\mathcal{W}$
	\ENDLOOP
	\LOOP[{{\color{gray}~~~\#  Stop after $K$ inner loops}}]
	\STATE Fit a set of GMM having 2 to $k_{max}$ kernels on $\mathcal{W}$
	\STATE Select the GMM with best Akaike Information Criterion
	\LOOP[{$N$ times}]
	    \STATE $\rho_{rnd} \%$ of the time, sample a random parameter $p \in \mathcal{P}$
	    \STATE Else, sample $p$ from a Gaussian chosen proportionally to its mean ALP value 
		\STATE Send $E(a \sim \mathcal{A}(p))$ to student $s_\theta$ and observe episodic reward $r_p$
	    \STATE Compute ALP of $p$ based on $r_p$ and $H$
	    \STATE Store $(p,r_p)$ pair in $H$, store $(p,ALP_p)$ pair in $\mathcal{W}$
	\ENDLOOP
	\ENDLOOP
	\STATE \textbf{Return} $s_\theta$
	
	\end{algorithmic}
\end{algorithm*}

\section{AGAIN}
\label{app-again}
\paragraph{IN variants.} In order to filter the list $\mathcal{C}_{raw}$ (see eq. \ref{eq:craw}) of GMMs extracted from a training trajectory $\tau_{s}$ selected in training trajectory history $\mathcal{H}$ into $\mathcal{C}$ and use it as an expert curriculum, we remove any Gaussian with a $LP_{ti}$ below $\delta_{LP}=0.2$ (the LP dimension is normalized between $0$ and $1$, which requires to choose an approximate potential reward range, set to $[-150,350]$ for all experiments on Box2D locomotion environments (sec. \ref{sec:exp:walker-climber} and sec. \ref{exp:trying-again}). When all Gaussians of a GMM are discarded, the GMM is removed from $\mathcal{C}$. In practice, it allows  to 1) remove non-informative GMMs corresponding to the initial exploration phase of ALP-GMM, when the learner has not made any progress (hence no LP detected by the teacher), and 2) remove an entire training trajectory $\tau_{s}$ if ALP-GMM never detected high-LP Gaussians, i.e. it failed to train student $s$. $\mathcal{C}$ is then iterated over to generate a curricula with either of the Time-based (see algo. \ref{int-algo}), Pool-based (see algo \ref{inp-algo}) or Reward-based (The one used in our main experiments, see algo \ref{inr-algo}) IN. The IN-P approach does not require additional hyperparameters. The IN-T requires an update rate $N$ to iterate over $\mathcal{C}$, which we set to $250$ (same as the fitting rate of ALP-GMM). The IN-R approach requires to extract additional data from the first run, in the form of a list $\mathcal{R}_{raw}$:
\begin{equation}
\label{eq:rraw}
    \mathcal{R}_{raw} = \{\mu_r^1, ...,\mu_r^t, \mu_r^T\} ~~s.t~~~ |\mathcal{R}_{raw}| = |\mathcal{C}_{raw}|,
\end{equation}
with T the total number of GMMs in the first run (same as in $\mathcal{C}_{raw}$), and $\mu_r^t$ the mean episodic reward obtained by the first DRL agent during the last $50$ tasks sampled from the $t^{th}$ GMM. $\mathcal{R}$ is simply obtained by removing any $\mu_r^t$ that corresponds to a GMM discarded while extracting $\mathcal{C}$ from $\mathcal{C}_{raw}$. The remaining rewards are then used as thresholds in IN-R to decide when to switch to the next GMM in $\mathcal{C}$.

\paragraph{AGAIN} In AGAIN (see algo. \ref{again-algo}), the idea is to use both IN (R,T or P) and ALP-GMM (without the random bootstrapping period) for curriculum generation. Our main experiments use IN-R as it is the highest performing variant (see app. \ref{ann:toy-exp}). This means that in the main sections of this paper, AGAIN $=$ AGAIN-R and IN $=$ IN-R. We combine the changing GMM of IN and ALP-GMM over time, simply by building a GMM $G$ containing Gaussians from the current GMM of IN and ALP-GMM. By selecting the Gaussian in $G$ from which to sample a new task using their respective LP, this approach allows to adaptively modulate the task sampling between both, shifting the sampling towards IN when ALP-GMM does not detect high-LP subspaces and towards ALP-GMM when the current GMM of IN have lower-LP Gaussians. While combining ALP-GMM to IN, we reduce the residual random sampling of ALP-GMM from $\rho_{high}=10\%$, used for the pretrain phase, to either $\rho_{low}=2\%$ for experiments presented in sec. \ref{sec:exp:toy-env} and sec. \ref{exp:trying-again}, or $\rho_{low}=0\%$  for experiments done in the Parkour environment in sec. \ref{sec:exp:walker-climber} (here we found $\rho_{low}=0\%$ to be beneficial in terms of performances w.r.t. $\rho_{low}=2\%$, which means that the task-exploration induced by the periodic GMM fit of ALP-GMM was sufficient for exploration). In AGAIN-R and AGAIN-T, when the last GMM $p(T)$ of the IN curriculum is reached, we switch the fixed $LP_{Ti}$ values of all IN Gaussians to periodically updated LP estimates, i.e. we allow AGAIN to modulate the importance of $p(T)$ for task sampling depending on its current student's performance.

\begin{algorithm*}[htb!]
	\caption{~ Pretrain phase (helper function)}
	\label{pret-algo}
	\begin{algorithmic}[1]
	\REQUIRE Student policy $s_\theta$, teacher training history $\mathcal{H}$, task-encoding parameter space $\mathcal{P}$, LP threshold $\delta_{LP}$, experimental pre-train budget $K_{pre}$, pre-test set size $m$, number of neighbors for student selection $k$, random sampling ratio $\rho_{high}$, parametric procedural environment generator $E$
	\vspace{0.2cm}
	\STATE Init $s_{\theta}$, train it for $K_{pre}$ env. steps with ALP-GMM($\rho_{high}, \mathcal{P}$) 
	\STATE Pre-test $s_{\theta}$ with $m$ tasks selected uniformly over $\mathcal{P}$ and get $KC_s^{pre}$ \COMMENT{{\color{gray}~~~\# Pre-test phase}}
	\STATE Apply knn algorithm in KC space of $\mathcal{H}$, get $k$ students closest to $KC_s^{pre}$ \STATE Among those $k$, keep the one with highest summed post training $KC^{post}$, extract its $\mathcal{C}_{raw}$
	\STATE Get $\mathcal{C}$ from $\mathcal{C}_{raw}$ by removing any Gaussian with $LP_{ti} < \delta_{LP}$.
	\STATE \textbf{Return} $\mathcal{C}$
	\end{algorithmic}
\end{algorithm*}

\begin{algorithm*}[htb!]
	\caption{~ Inferred progress Niches - Time-based (IN-T)}
	\label{int-algo}
	\begin{algorithmic}[1]
	
	\REQUIRE Student policy $s_\theta$, teacher training history $\mathcal{H}$, task-encoding parameter space $\mathcal{P}$, LP threshold $\delta_{LP}$, update rate $N$, experimental budget $K$, experimental pre-train budget $K_{pre}$, pre-test set size $m$, number of neighbors for student selection $k$, random sampling ratio $\rho_{high}$, parametric procedural environment generator $E$
	\vspace{0.2cm}
	\STATE Launch Pretrain phase and get expert GMM list $\mathcal{C}$ \COMMENT{{\color{gray}~~~\# See algo. \ref{pret-algo}}}
	\STATE Initialize expert curriculum index $i_{c}$ to $0$
	\LOOP[{~Stop after $K - K_{pre}$ environment steps}]
	    \STATE Set $i_{c}$ to $min(i_{c}+1, len(\mathcal{C}))$
	    \STATE Set current GMM $G_{IN}$ to $i_{c}^{th}$ GMM in $\mathcal{C}$
	    \LOOP[{$N$ times}]
	    \STATE Sample $p$ from a Gaussian in $G_{IN}$ chosen proportionally to its $LP_{ti}$
		\STATE Send $E(a \sim \mathcal{A}(p))$ to student $s_{\theta}$
	\ENDLOOP
	\ENDLOOP
	\STATE Add student's training trajectory to $\mathcal{H}$
	\STATE \textbf{Return} $s_\theta$
	\end{algorithmic}
\end{algorithm*}

\begin{algorithm*}[htb!]
	\caption{~ Inferred progress Niches - Pool-based (IN-P)}
	\label{inp-algo}
	\begin{algorithmic}[1]
	
	\REQUIRE Student policy $s_\theta$, teacher training history $\mathcal{H}$, task-encoding parameter space $\mathcal{P}$, LP threshold $\delta_{LP}$, update rate $N$, experimental budget $K$, experimental pre-train budget $K_{pre}$, pre-test set size $m$, number of neighbors for student selection $k$, random sampling ratio $\rho_{high}$, parametric procedural environment generator $E$
	\vspace{0.2cm}
	\STATE Launch Pretrain phase and get expert GMM list $\mathcal{C}$ \COMMENT{{\color{gray}~~~\# See algo. \ref{pret-algo}}} 
	\STATE Initialize pool GMM $G_{IN}$, containing all Gaussians from $\mathcal{C}$
	\LOOP[{~Stop after $K - K_{pre}$ environment steps}]
	    \STATE Sample $p$ from a Gaussian in $G_{IN}$ chosen proportionally to its $LP_{ti}$
		\STATE Send $E(a \sim \mathcal{A}(p))$ to student $s_\theta$
	\ENDLOOP
	\STATE Add student's training trajectory to $\mathcal{H}$
	\STATE \textbf{Return} $s_\theta$
	\end{algorithmic}
\end{algorithm*}

\begin{algorithm*}[htb!]
	\caption{~ Inferred progress Niches - Reward-based (IN-R)}
	\label{inr-algo}
	\begin{algorithmic}[1]
	
	\REQUIRE Student policy $s_\theta$, teacher training history $\mathcal{H}$, task-encoding parameter space $\mathcal{P}$, LP threshold $\delta_{LP}$, update rate $N$, experimental budget $K$, experimental pre-train budget $K_{pre}$, pre-test set size $m$, number of neighbors for student selection $k$, random sampling ratio $\rho_{high}$, parametric procedural environment generator $E$
	\vspace{0.2cm}
	\STATE Launch Pretrain phase and get expert GMM list $\mathcal{C}$ \COMMENT{{\color{gray}~~~\# See algo. \ref{pret-algo}}} 
	\STATE Initialize reward First-in-First-Out window $\mathcal{W}$, set max size to $N$
	\STATE Initialize expert curriculum index $i_{c}$ to $0$
	\LOOP[{~Stop after $K - K_{pre}$ environment steps}]
	    \STATE If $\mathcal{W}$ is full, compute mean reward $\mu_w$ from $\mathcal{W}$
	    \STATE ~~~~If $\mu_w$ superior to $i_{c}^{th}$ reward threshold in $\mathcal{R}$, set $i_{c}$ to $min(i_{c}+1, len(\mathcal{C}))$
	    \STATE Set current GMM $G_{IN}$ to $i_{c}^{th}$ GMM in $\mathcal{C}$
	    \STATE Sample $p$ from a Gaussian in $G_{IN}$ chosen proportionally to its $LP_{ti}$
		\STATE Send $E(a \sim \mathcal{A}(p))$ to student $s_\theta$ and add episodic reward $r_p$ to $\mathcal{W}$
	\ENDLOOP
	\STATE Add student's training trajectory to $\mathcal{H}$
	\STATE \textbf{Return} $s_\theta$
	\end{algorithmic}
\end{algorithm*}

\begin{algorithm*}[htb!]
	\caption{~ Alp-Gmm And Inferred progress Niches (AGAIN)}
	\label{again-algo}
	\begin{algorithmic}[1]
	
	\REQUIRE Student policy $s_\theta$, teacher training history $\mathcal{H}$, task-encoding parameter space $\mathcal{P}$, LP threshold $\delta_{LP}$, update rate $N$, experimental budget $K$, experimental pre-train budget $K_{pre}$, pre-test set size $m$, number of neighbors for student selection $k$, random sampling ratio $\rho_{low}$ and $\rho_{high}$, parametric procedural environment generator $E$
	\vspace{0.2cm}
	\STATE Launch Pretrain phase and get expert GMM list $\mathcal{C}$ \COMMENT{{\color{gray}~~~\# See algo. \ref{pret-algo}}}
	\STATE Setup new ALP-GMM($\rho_{rnd}=0, \mathcal{P}$) \COMMENT{{\color{gray}~~~\# See algo. \ref{algo:ALP-GMM}}} 
	\STATE Setup either IN-T, IN-P or IN-R \COMMENT{{\color{gray}~~~\# See algo. \ref{int-algo}, \ref{inp-algo} and \ref{inr-algo}}} 
	\LOOP[{~Stop after $K - K_{pre}$ environment steps}]
	    \STATE Get composite GMM $G$ from the current GMM of both ALP-GMM and IN
	    \STATE $\rho_{low} \%$ of the time, sample a random parameter $p \in \mathcal{P}$
	    \STATE Else, sample $p$ from a Gaussian chosen proportionally to its $LP$ 
		\STATE Send $E(a \sim \mathcal{A}(p))$ to student $s_\theta$ and observe episodic reward $r_p$
	    \STATE Send $(p,r_p)$ pair to both ALP-GMM and IN
	\ENDLOOP
	\STATE Add student's training trajectory to $\mathcal{H}$
	\STATE \textbf{Return} $s_\theta$
	
	\end{algorithmic}
\end{algorithm*}

\section{Considered ACL and Meta-ACL teachers}
\label{an:details}
\paragraph{Meta-ACL variants} Our proposed approach, AGAIN, is based on the combination of an inferred expert curriculum with ALP-GMM, an exploratory ACL approach. In section \ref{sec:methods} and appendix \ref{app-again}, we present $3$ approaches to use such an expert curriculum, giving the AGAIN-R, AGAIN-P and AGAIN-T algorithms. In our experiments, we also consider ablations were we only use the expert curriculum, giving the IN-R, IN-P and IN-T variants. We also consider two additional AGAIN variants that do not use our proposed KC-based student selection method:
\begin{itemize}
    \item AGAIN with Random selection (AGAIN\_RND), a lower-baseline ablation were we select the training trajectory $\tau$ from which to extract the expert curriculum randomly in history $\mathcal{H}$.
    \item AGAIN with Ground Truth selection (AGAIN\_GT), an upper-baseline using privileged information. Instead of performing the knn algorithm in the KC space, this approach directly uses the true student distribution. For instance, in the Parkour environment, given a new student $s$, AGAIN\_GT selects the $k$ previously trained students from $\mathcal{H}$ that are morphologically closest to $s$ (i.e. same embodiment type and closest limb sizes), and uses the training trajectory of the student with highest score $j_s$ (see sec. \ref{sec:methods}).
    
\end{itemize}
Note that both for AGAIN\_RND and AGAIN\_GT, there is no need to pre-test the student, which means we can use the IN expert curriculum directly at the beginning of training rather than after a pre-training phase.


\paragraph{ACL conditions} A first natural ACL approach to compare our AGAIN variants to is ALP-GMM, the underlying ACL algorithm in AGAIN. We also add as a lower-baseline a random curriculum teacher (Random), which samples tasks' parameters randomly over the task space.

In both the toy environment (sec. \ref{sec:exp:toy-env}, toy env. for short) and the Parkour environment (sec. \ref{sec:exp:walker-climber}), we additionally compare to Adaptive Domain Randomization (ADR), an ACL algorithm proposed in \cite{OpenAI2019SolvingRC}, which is based on inflating a task distribution sampling from a predefined initially feasible task $p_{easy}$ (w.r.t a given student). Each lower and upper boundaries of each dimension of the sampling distribution are modified independently with step size $\Delta_{step}$ whenever a predefined mean reward threshold $r_{thr}$ is surpassed over a window (of size $q$) of tasks occasionally sampled (with probability $\rho_b$) at the sampling dimension boundary. More details can be found in \cite{OpenAI2019SolvingRC}. In our experiments, as we do not assume access to expert knowledge over students sampled within the student distribution, we randomize the setting of $p_{easy}$ uniformly over the task space in Parkour experiments and uniformly over the $4$ possible student starting subspaces in toy env. experiments. Based on the hyperparameters proposed in \cite{OpenAI2019SolvingRC} and on informal hyperparameter search, we use $[\rho_b=0.5, r_{thr}=1, \Delta_{step}=0.05, q=10]$ in toy env. experiments and $[\rho_b=0.5, r_{thr}=230, \Delta_{step}=0.1, q=20]$ in Parkour experiments.

In experiments described in sec \ref{exp:trying-again}, we compare our approaches to an oracle condition (Oracle), which is a hand-made curriculum that is very similar to IN-R, except that the list $\mathcal{C}$ is built using expert knowledge before training starts (i.e. no pre-train and pre-test phases), and all reward thresholds $\mu_r^i$ in $\mathcal{R}$ (see eq. \ref{eq:rraw}) are set to $230$, which is an episodic reward value often used in the literature as characterizing a default walker having a "reasonably efficient" walking gate in environments derived from the Box2D gym environment BipedalWalker \cite{poet,portelas2019}.In practice, Oracle starts proposing tasks from a Gaussian (with std of $0.05$) located at the simplest subspace of the task space (ie. low stump height and high stump spacing) and then gradually moves the Gaussian towards the hardest subspaces (high stump height and low stump spacing) by small increments ($50$ steps overall) happening whenever the mean episodic reward of the DRL agent over the last $50$ proposed tasks is superior to $230$.

\newpage\section{Analysing Meta-ACL in a toy environment}
\label{ann:toy-exp}

\begin{figure*}[htb!]
    \centering
    \subfloat{\includegraphics[width=0.3\textwidth]{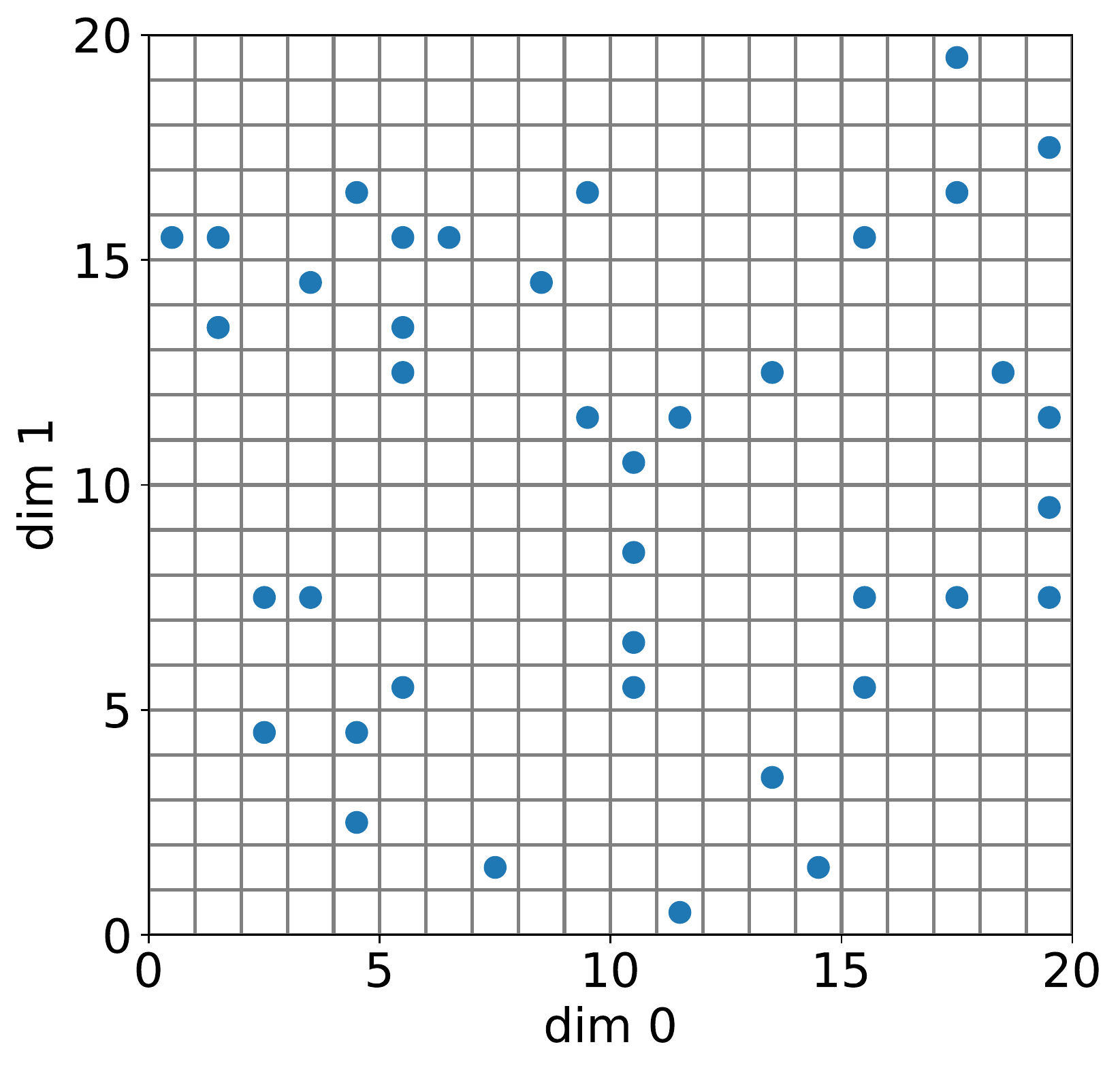}}
    \subfloat{\includegraphics[width=0.3\textwidth]{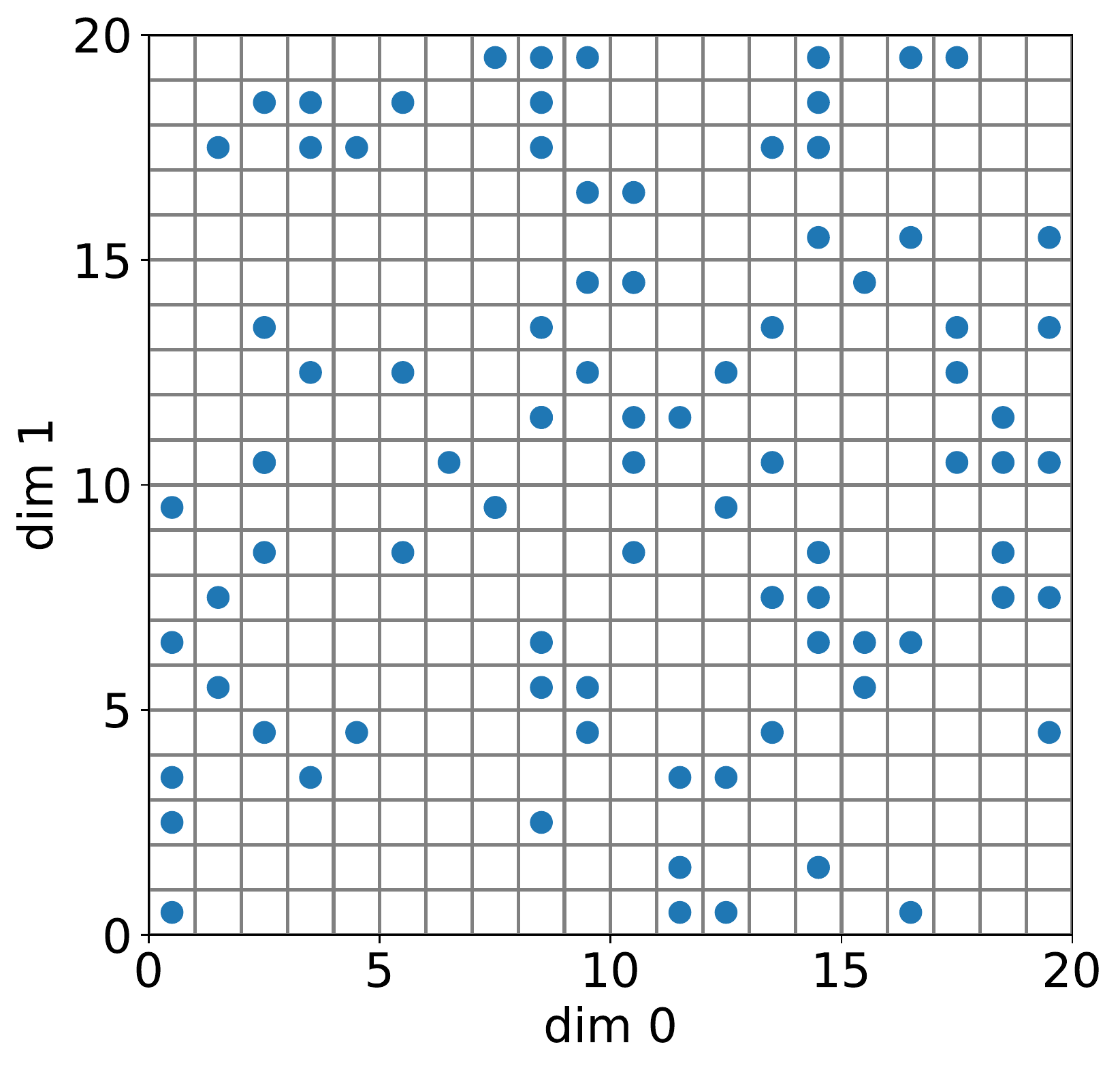}}
    \caption{\footnotesize{ Additional visualizations for the varying classroom size experiments (see sec. \ref{sec:exp:toy-env}).
    \textbf{Left:} Visualization of the starting cells of students from a $10$\% sample of a classroom of $400$ students (one per student type) trained with ALP-GMM and used to populate the training trajectory $\mathcal{H}$. Each blue circle marks the starting cell of each student (i.e. its type) within the $2$D parameter space $\mathcal{P}$, which is an initial learning subspace that needs to be detected by the teacher for successful training.  \textbf{Right:}Visualization of the fixed set of $96$ randomly drawn students that have to be trained by Meta-ACL variants given $\mathcal{H}$. As not all student types are represented in $\mathcal{H}$, Meta-ACL approaches have to generalize their curriculum generation to these new students.}}
    \label{classroom-toyenv-vizu}
\end{figure*}

\begin{figure*}[htb!]
\centering
\subfloat[with type 0 new student]{\includegraphics[width=0.35\textwidth]{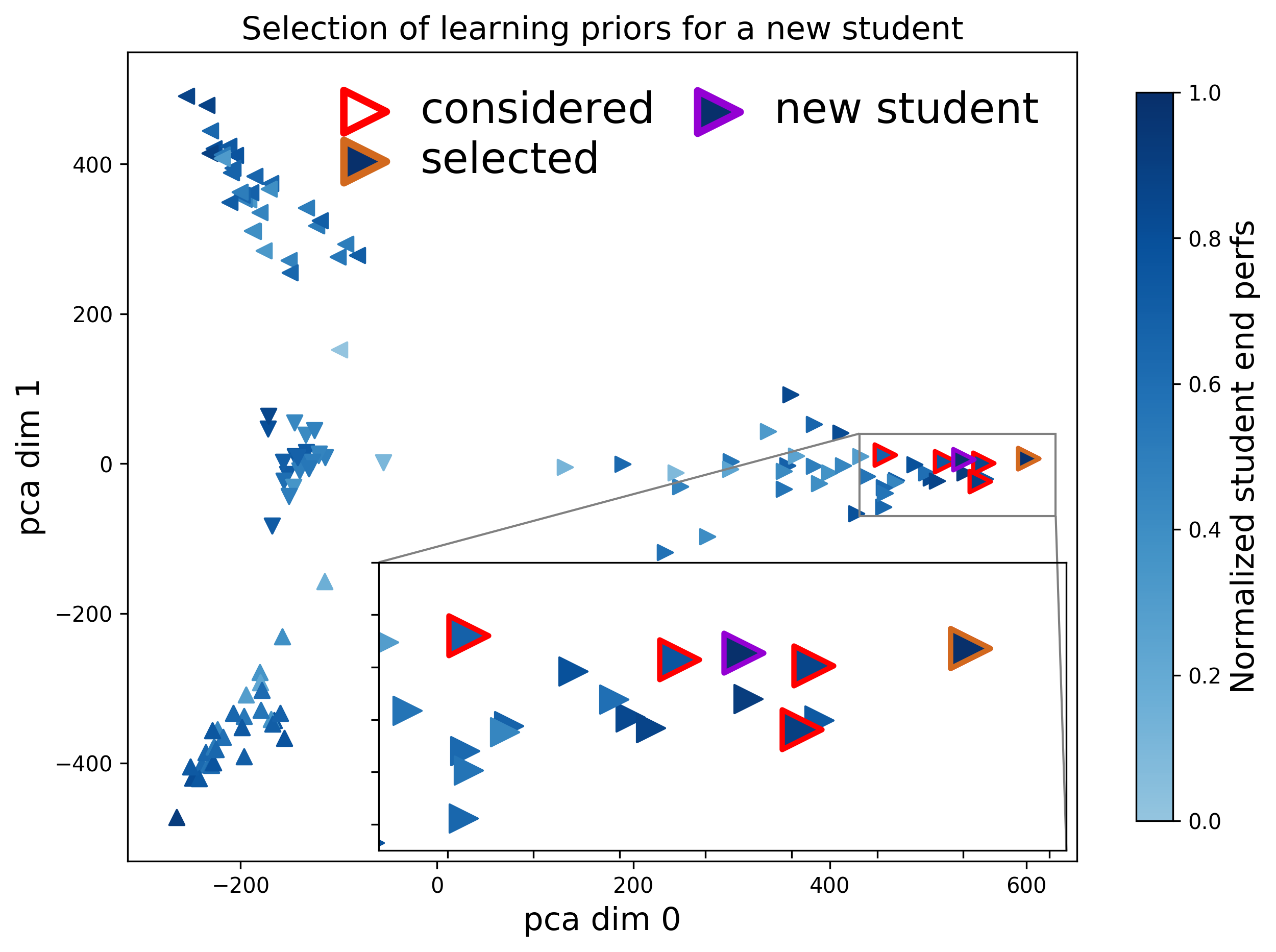}}
\subfloat[with type 1 new student]{\includegraphics[width=0.35\textwidth]{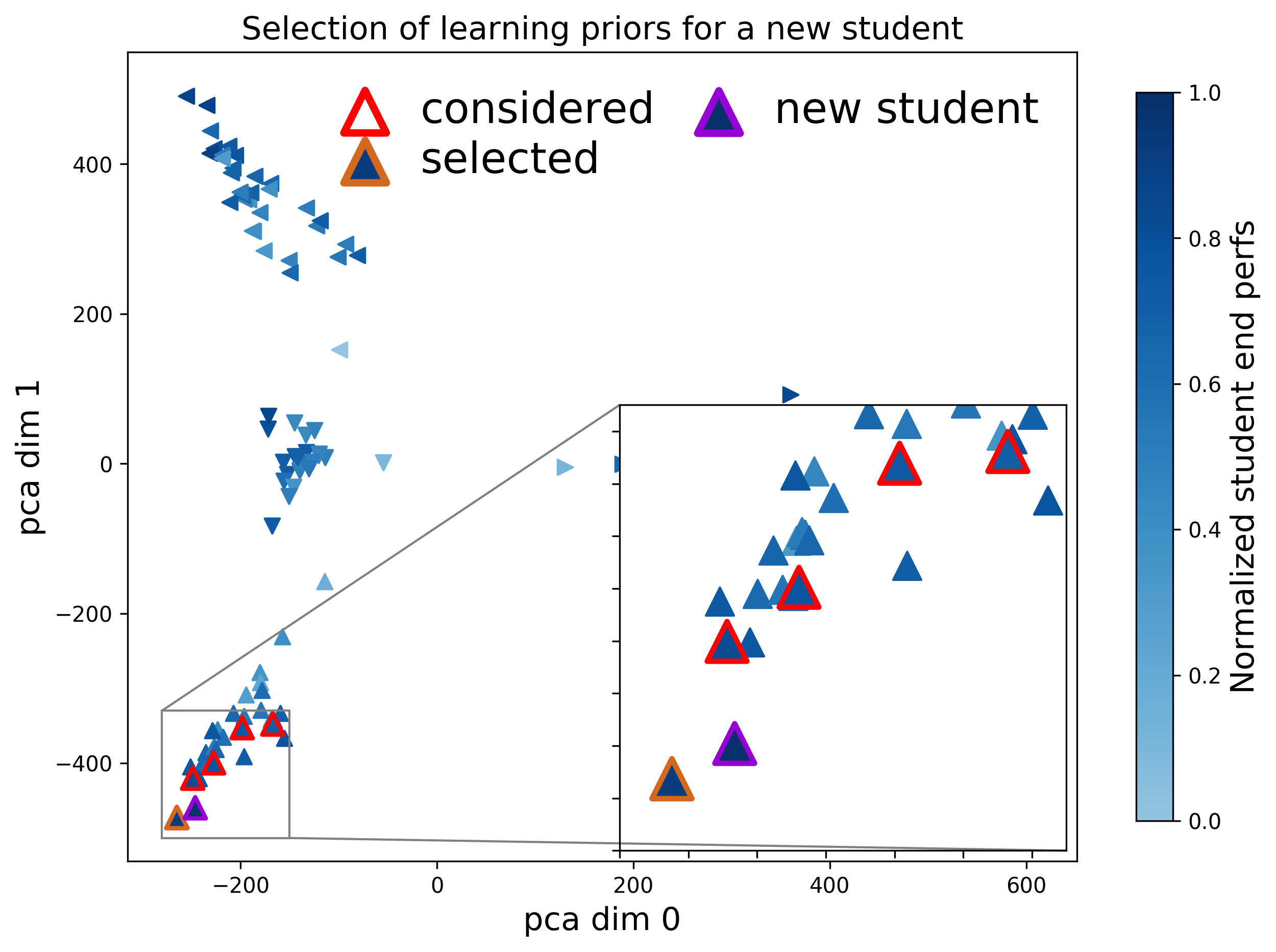}}

\subfloat[with type 2 new student]{\includegraphics[width=0.35\textwidth]{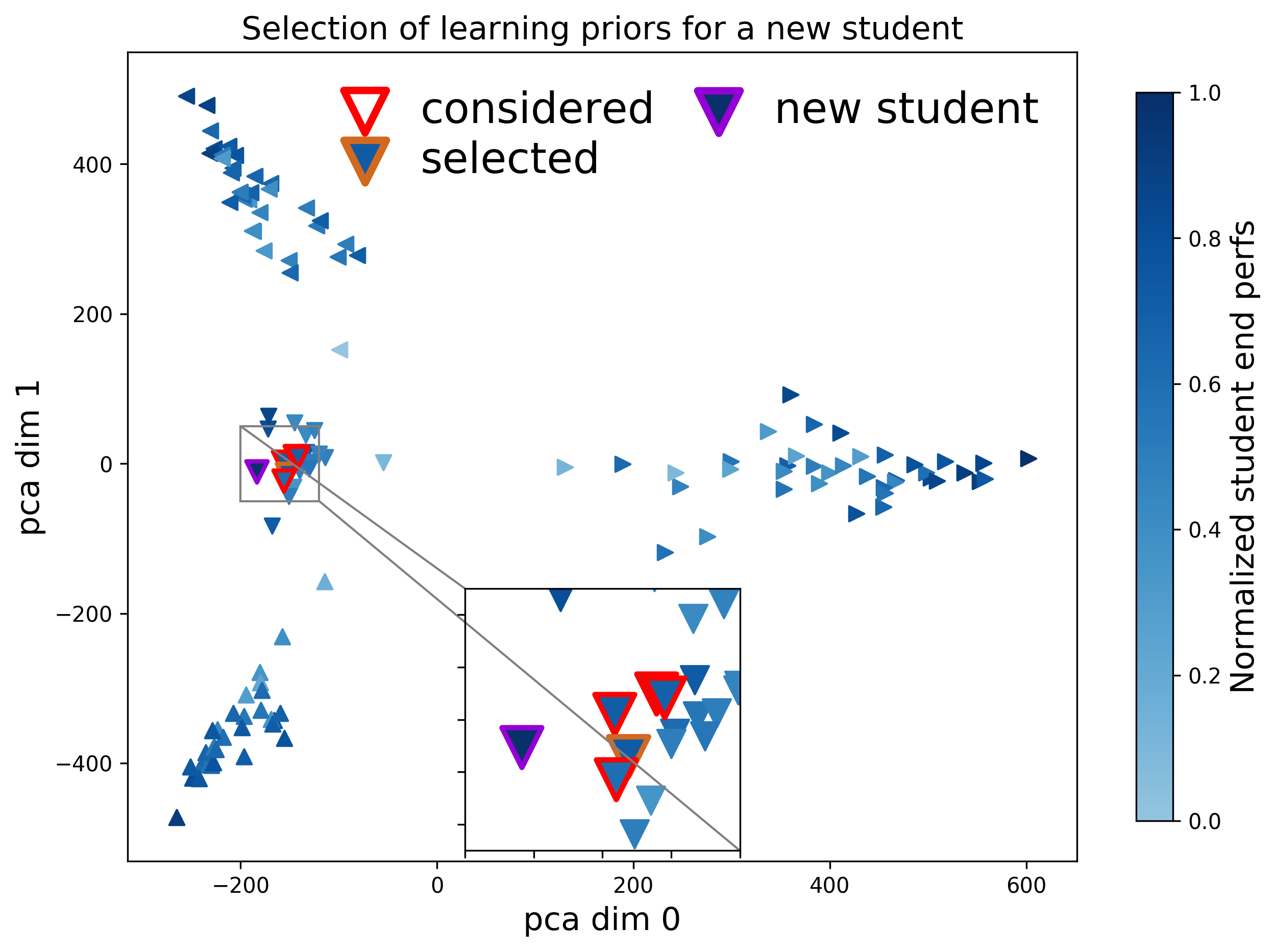}}
\subfloat[with type 3 new student]{\includegraphics[width=0.35\textwidth]{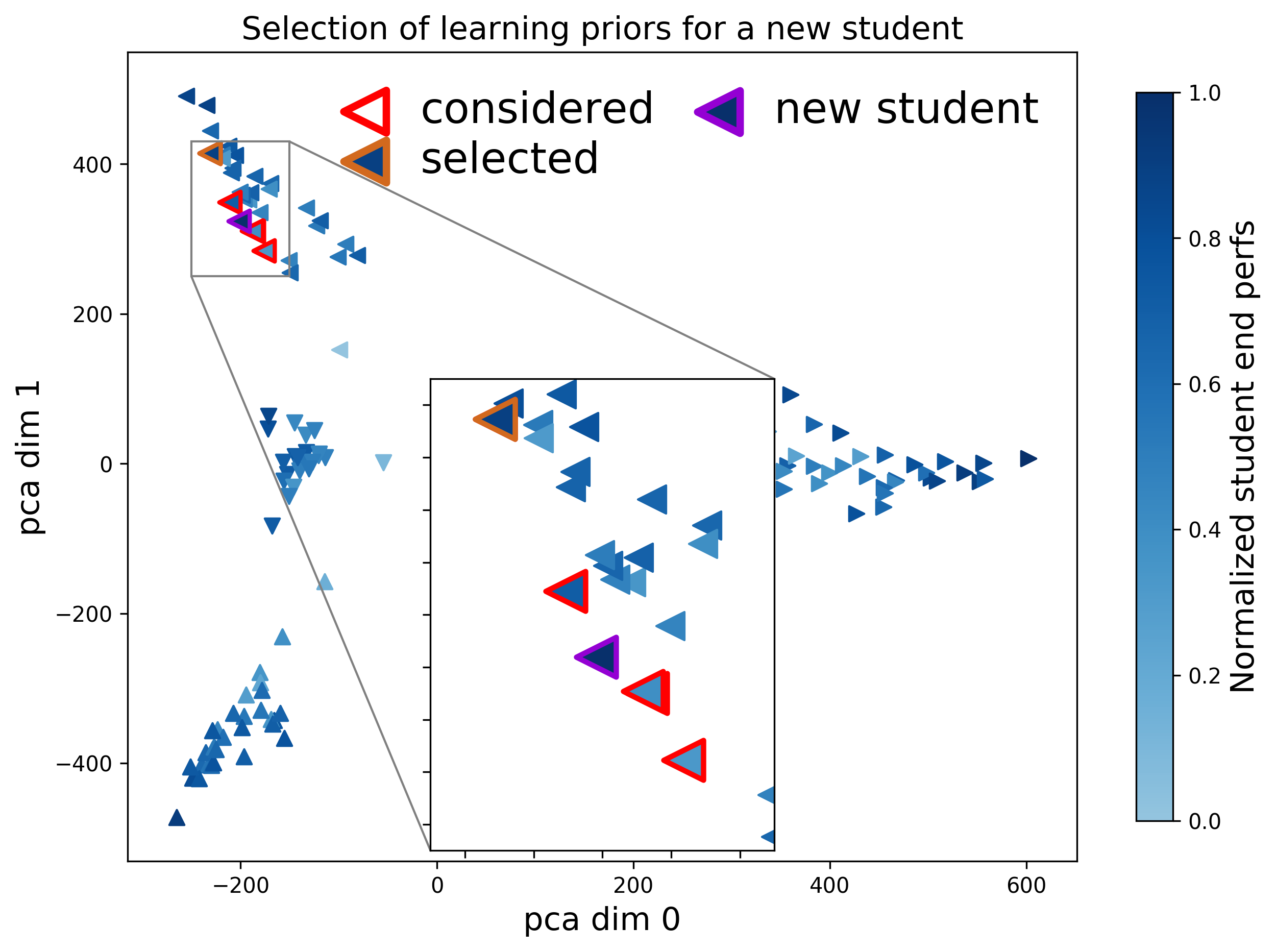}}

\caption{\footnotesize{\textbf{Examples of student selection process in $4$-student type toy environment.} In all figures, we plot the $2$D PCA visualization of the $KC^{pre}$ vectors (after pre-training) of the initial classroom ($128$ students) trained with ALP-GMM and used to populate the training trajectory $\mathcal{H}$ used by AGAIN variants in our $4$-student type toy env experiments (see sec. \ref{sec:exp:toy-env}). We then use these $4$ figures to showcase the selection process happening in $4$ different AGAIN-R runs (one per student type). Each triangle represents a student, whose ground truth type (i.e. its initial learning cell) is denoted by the orientation of the triangle. Given a new student to train, AGAIN pretrains the student, constructs its KC vector (purple border triangle), infers the k closest previously trained students from $\mathcal{H}$ (red and golden border triangles), and use the one with highest end of training performance (i.e. highest score $s$, see sec. \ref{sec:methods}), denoted by a golden border triangle, to infer curriculum priors for the new student.}}
\label{toy-exps-stud-selec-vizu}
\end{figure*}

In this section we report the full comparative experiments done in the toy environment, which includes comparisons with AGAIN-T and AGAIN-P to AGAIN-R, shown in table \ref{toy-env-results-table}. We also provide visualizations of the KC-based curriculum priors selection process (see fig. \ref{toy-exps-stud-selec-vizu}) happening after the pretraining phase in AGAIN along with a visualization of the fixed set of $96$ randomly drawn students used to perform the varying classroom experiments reported in sec. \ref{sec:exp:toy-env} (see fig. \ref{classroom-toyenv-vizu}).

\paragraph{Additional comparative analysis} Table \ref{toy-env-results-table} summarizes the post-training performances obtained by our considered Meta-ACL conditions and ACL baselines on the toy environment with only $4$ possible students on a fixed set of $48$ randomly drawn students. Meta-ACL conditions are given a training trajectory $\mathcal{H}$ created by training an initial classroom of $128$ students. Using a Reward-based iterating scheme over the inferred expert curriculum (AGAIN-R and IN-R) outperforms the Time-based and Pool-based variants ($p<.001$). This result was expected as both these last two variants do not have flexible mechanisms to adapt to the student being trained. The pool based variants (AGAIN-P and IN-P), which discard the temporal ordering of the expert curriculum are the worst performing variants, statistically significantly inferior to both Reward-based and Time-based conditions ($p<.001$).

\begin{table}[htb!]
\caption{\footnotesize{\textbf{Experiments on the toy environment.} The average performance with standard deviation after 200k episodes is reported (48 seeds per conditions). For Meta-ACL variants we report results with column 1) the regular KC-based curriculum prior selection performed after $20$k pre-training episodes, column 2) An ablation that performs the selection at random before training, and column 3) An oracle condition selecting before training the curriculum prior using student ground truth type. \textit{*} Denotes stat. significant advantage w.r.t. ALP-GMM (Welch's t-test at $200k$ ep. with $p<0.05$).}}
\vspace{0.3cm}
    \centering
    \footnotesize
\begin{tabular}{@{}llll@{}}
\toprule
Condition & Regular        & Random  & Ground Truth  \\ \midrule
AGAIN-R   & 98.8 +- 4.8* & 55.4 +- 32.2     & 99.8 +- 0.9*         \\ 
IN-R      & 91.4 +- 3.4* & 26.3 +- 41.1     & 92.5 +- 3.0*         \\ 
AGAIN-T   & 84.3 +- 3.8    & 38.6 +- 34.1     & 89.0 +- 1.7*         \\ 
IN-T      & 79.0 +- 12.0   & 30.3 +- 37.3     & 88.9 +- 1.7*         \\ 
AGAIN-P   & 38.2 +- 7.5    & 9.3 +- 9.2      & 14.8 +- 1.2            \\ 
IN-P      & 40.6 +- 6.4    & 9.2 +- 9.0       & 15.1 +- 1.2            \\ \midrule
ALP-GMM   & 84.6 +- 3.4    &                  &                        \\ 
ADR       & 14.9 +- 27.4   &                  &                        \\ 
Random    & 10.0 +- 0.8    &                  &                        \\  \bottomrule
\end{tabular}

    \label{toy-env-results-table}
\end{table}

\begin{figure*}[b]
\centering
\subfloat{\includegraphics[width=0.35\textwidth]{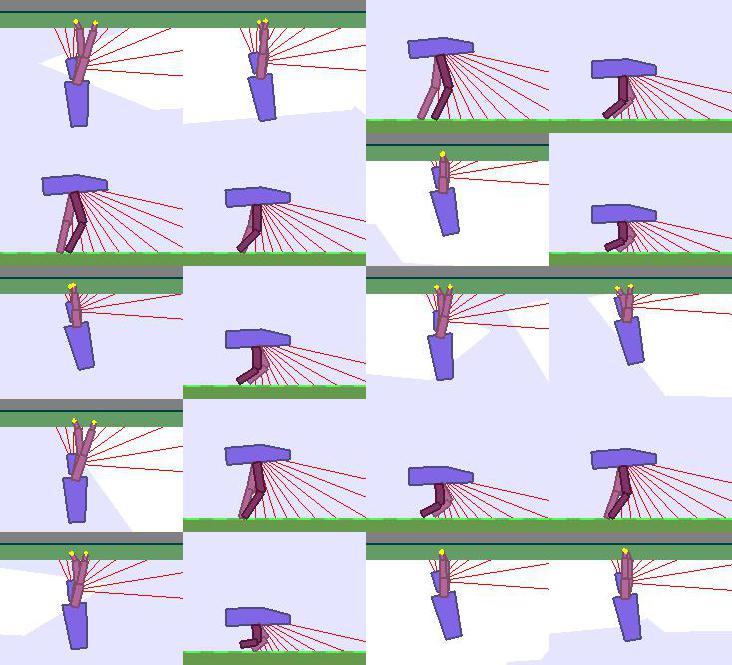}}\hspace{0.2cm}\subfloat{\includegraphics[width=0.35\textwidth]{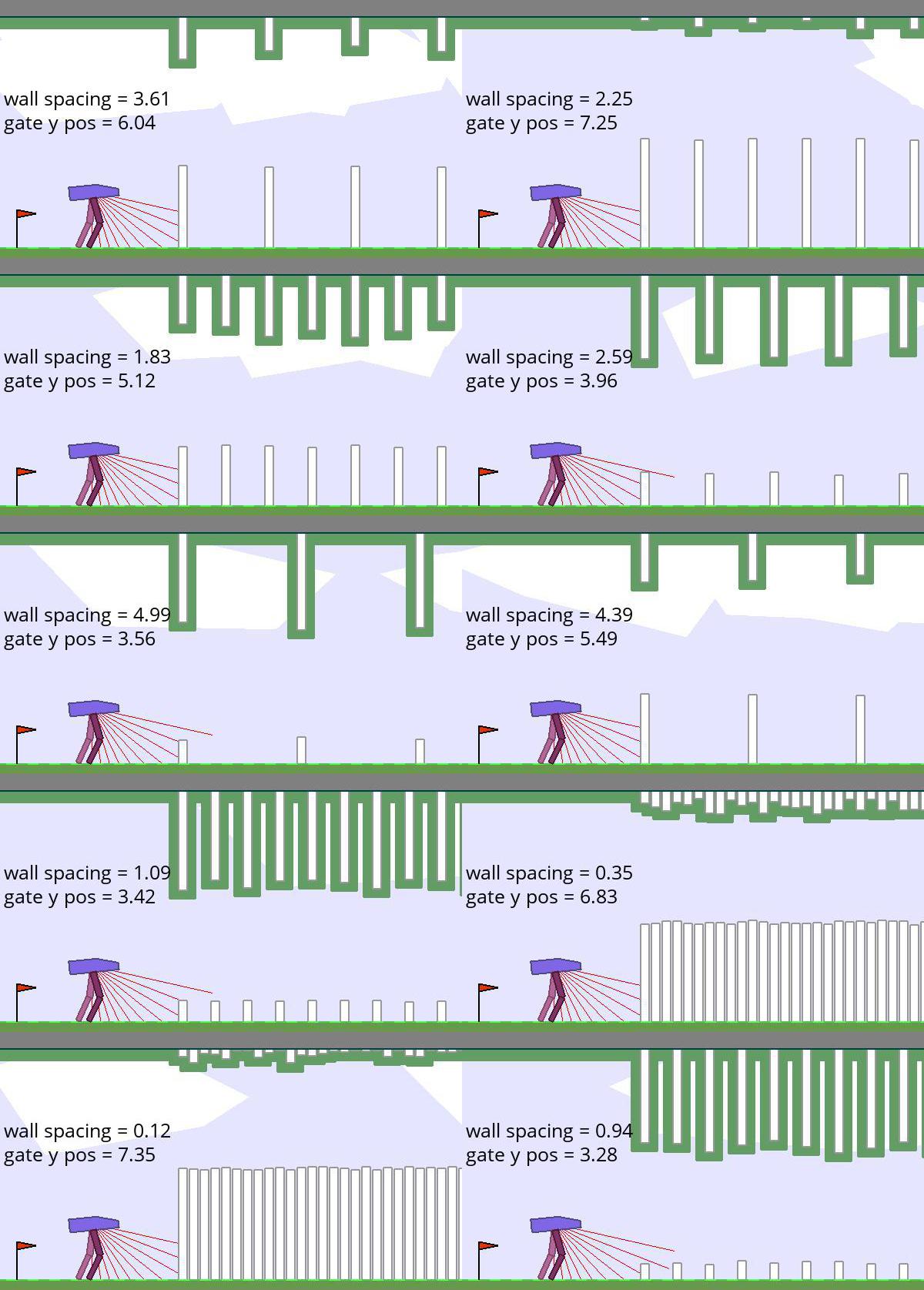}}
\caption{\footnotesize{Visualizations of the student space and the task space of the Parkour environment.\textbf{Left:} Examples of possible agent embodiments (randomly set for a given DRL learner before training starts). \textbf{Right:} Examples of randomly sampled parkour tracks.}}
\label{parkour-env-vizu}
\end{figure*}

\vspace{10cm}\section{Meta-ACL for DRL students in the Parkour environment}
\label{ann:parkour}
In this section we give additional details on the Parkour environment presented in section \ref{sec:exp:walker-climber}, and we provide additional details and visualizations on the experiments that were performed on it.

\paragraph{Details on the Parkour environment.} In our experiments, we bound the wall spacing dimension of the task space to $\Delta_{w}=[0,6]$, and the gate y position to $\mu_{gate}=[2.5,7.5]$. In practice, given a single parameter tuple $(\mu_{gate},\Delta_{w})$, we actually encode a distribution of tasks, since for each new wall along the track we add an independent Gaussian noise to each wall's gate y position $\mu_{gate}$. Examples of parkour tasks randomly sampled within these bounds are available in figure \ref{parkour-env-vizu} (right). At the beginning of training a given DRL policy, the agent is embodied in either a bipedal walker morphology with two joints per legs or a two-armed climber morphology with 3-joints per arms ended by a grasping "hand". Both morphologies are controlled by torque. Climbers have an additional action dimension $g \in [-1,1]$ used to grasp: if $g \in [-1,0[$, the climber closes its gripper, and if $g \in ]0,1]$ it keeps it open. To avoid falling (which aborts the episode with a $-100$ penalty) while moving forward to collect rewards, climber agents must learn to swing themselves forward by successive grasp-and-release action sequences. To increase the diversity of the student distribution, we also randomize limb sizes. See figure \ref{parkour-env-vizu} (left) for examples of randomly sampled embodiments.

\paragraph{Soft Actor-Critic students} In our experiments, we use an implementation of Soft Actor-Critic provided by OpenAI\footnote{https://github.com/openai/spinningup} (MIT license). We use a $2$ layered ($400$,$300$) network for V, Q1, Q2 and the policy. Gradient steps are performed each $10$ environment steps, with a learning rate of $0.001$ and a batch size of $1000$. The entropy coefficient is set to $0.005$.

\paragraph{Evaluation procedure} To report the performance of our students on the Parkour environment, we use two separate test sets, one per embodiment type. For walkers we use a $100$-tasks test set, uniformly sampled over a subspace of the task space with $\Delta_w \in [0,6]$ and $\mu_{gate} \in [2.5,3.6]$, which we chose based on 1) what we initially believed to be morphologically feasible for walkers, and 2) based on previously designed test sets built in recent work \cite{portelas2019} on comparable bipedal walker experiments). For climbers, because there is no similar experiments in the literature and since it is hard to infer beforehand what will be achievable by such a morphology, we simply use a uniform test set of $225$ tasks sampled over the full task space. Importantly, the customized test set used for walkers is solely used for visualization purposes. In our AGAIN approaches, we pre-test all students with the expert-knowledge-free set of $225$ tasks uniformly sampled over the task space. 

\paragraph{Compute resources} Each of the 576 seeds required to reproduce our experiments (128 seeds for the classroom and 7*64 seeds for our 7 conditions) takes 36 hours on a single cpu. This amounts to around 21 000 cpu hours. Each run requires less than 1GB of RAM.

\begin{figure*}[htb!]
\centering
\subfloat{\includegraphics[width=0.6\textwidth]{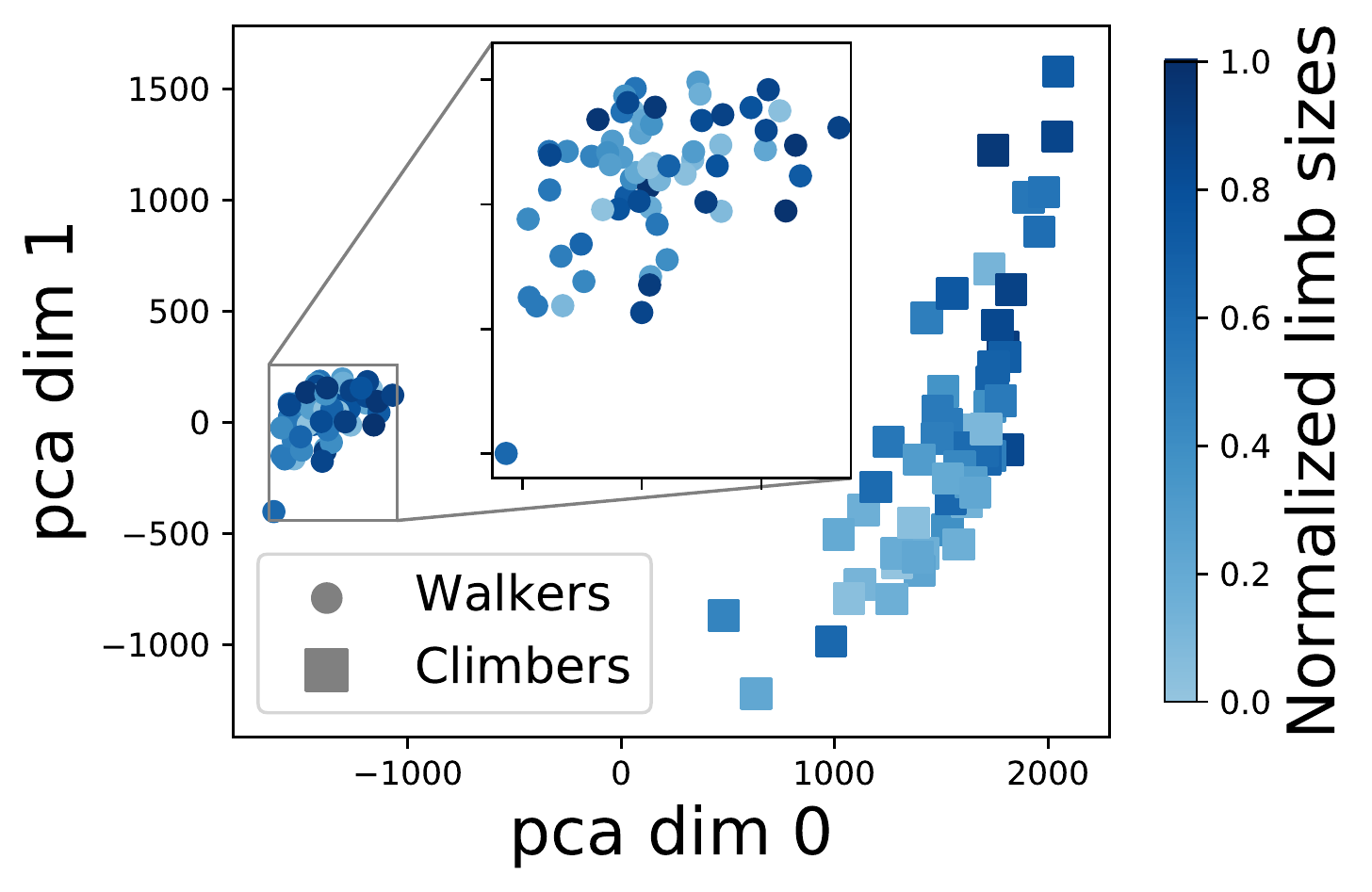}}

\subfloat{\includegraphics[width=0.45\textwidth]{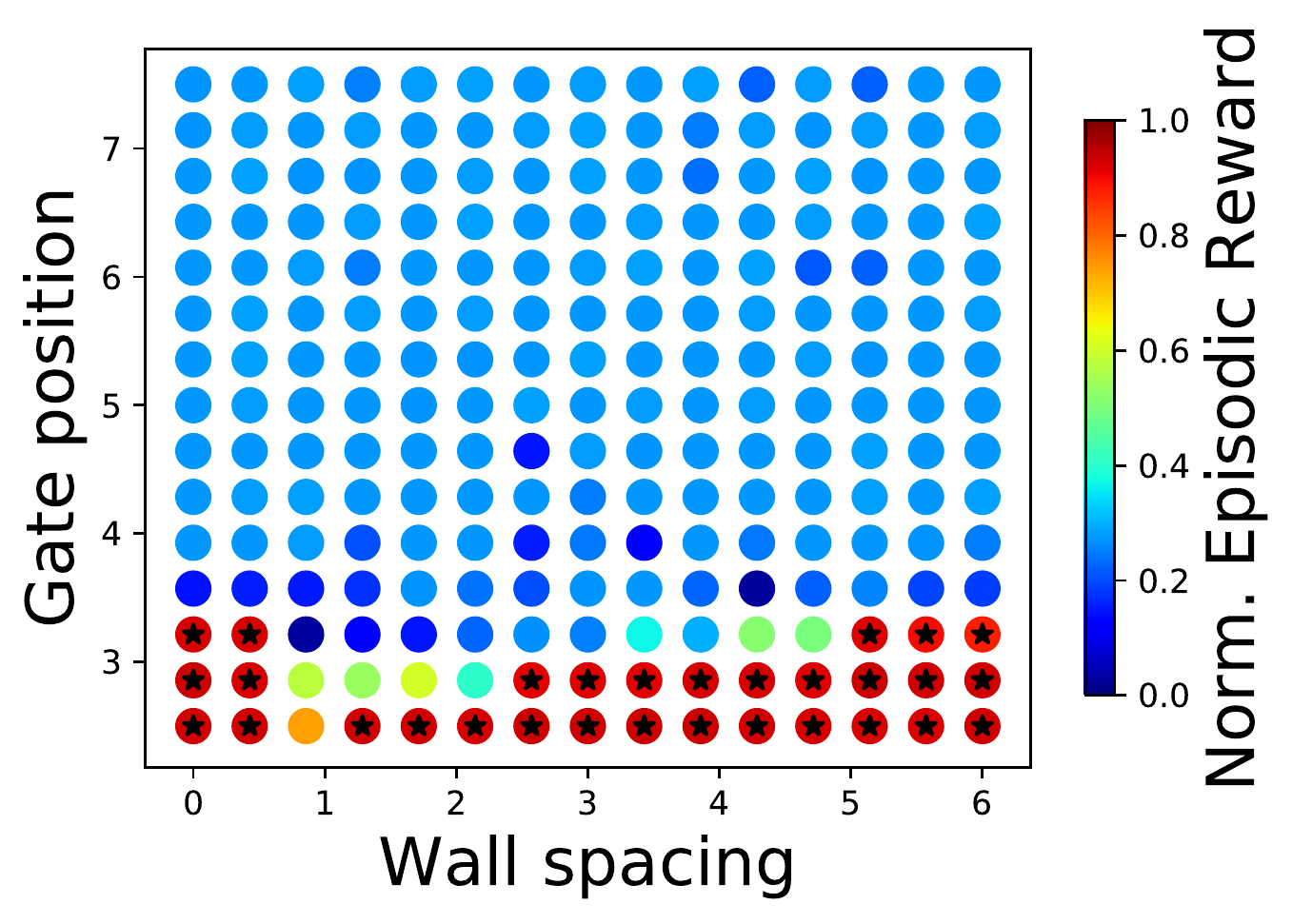}}\hspace{0.3cm}\subfloat{\includegraphics[width=0.45\textwidth]{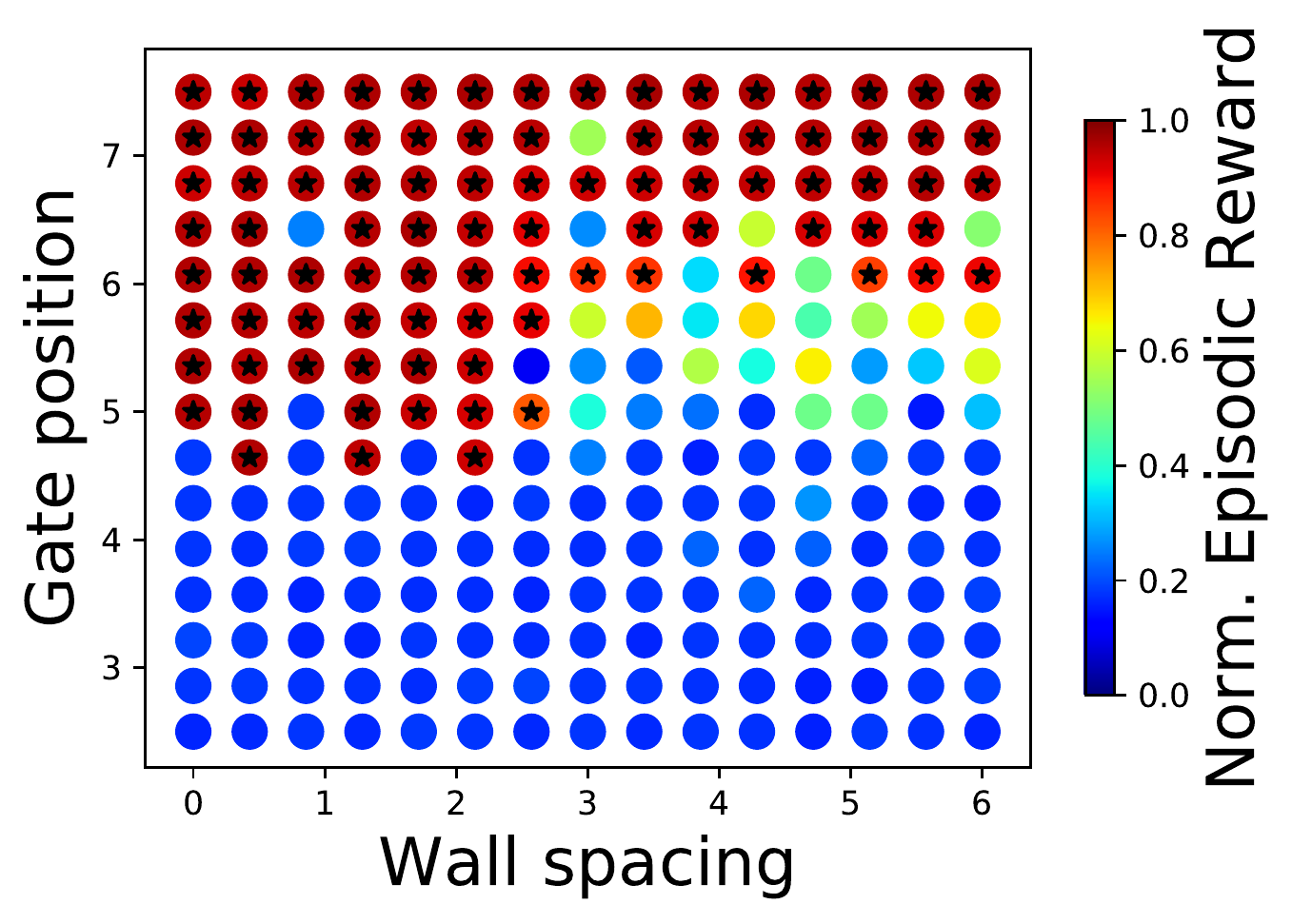}}
\caption{\footnotesize{\textbf{top:} PCA of classroom's KC vector (128 students) after being trained for 10M student steps with ALP-GMM. \textbf{left and right:} Episodic reward obtained for each task that compose the KC vector by a walker-student (left) and a climber-student (right) of this classroom. Stars are added for all tasks for which the agent obtained more than $r=230$ (which corresponds to an efficient locomotion policy). Walkers only manage to learn tasks with very low gate positions while climbers learn only tasks with medium to high gate positions.}}
\label{vizu-parkour-classroom}
\end{figure*}

\paragraph{Visualizing student diversity.} To assess whether our proposed multi-modal distribution of possible students in the Parkour environment do have diverse competence profiles (which is desirable as it creates a challenging Meta-ACL scenario), we plot the 2D PCA of the post training KC vector for each students of the initial classroom trained with ALP-GMM (used to populate $\mathcal{H}$). The result, visible in figure \ref{vizu-parkour-classroom} (top), shows that climber-students and walker-students are located in two independant clusters, i.e. they do have clearly different competence profiles. The spread of each clusters also demonstrates that variations in initial policy parameters and limb sizes also creates students with diverse learning potentials. The competence differences between walkers and climbers can also be seen in Figure \ref{vizu-parkour-classroom} (left and right), which shows the episodic reward obtained for each of the $225$ tasks of the KC vector after training by a representative walker student (left) and climber student (right).

\clearpage\newpage\section{Applying Meta-ACL to a single student: Trying AGAIN instead of trying longer}
\label{ann:tryingagain}

In the following section we report all experiments on applying AGAIN variants to train a single DRL student (i.e. no history $\mathcal{H}$), which is briefly presented in sec. \ref{exp:trying-again}.

\begin{figure}[htb!]

\centering\includegraphics[width=0.75\columnwidth]{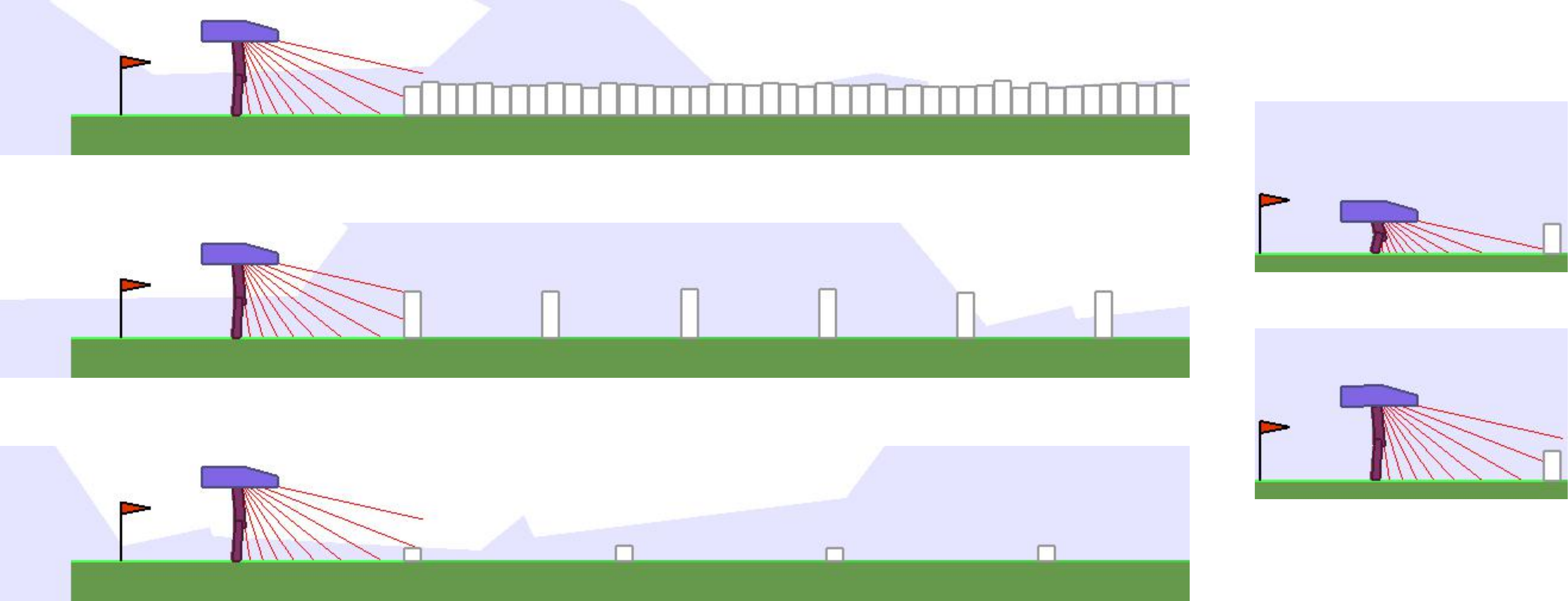}
\caption{\footnotesize{Parameterized BipedalWalker environment. \textbf{Left:} Examples of generated tracks. \textbf{Right:} The two walker morphologies tested on the environment.} One parameter tuple ($\mu_h, \delta_s$) actually encodes a \textit{distribution} of tasks as the height of each stump along the track is drawn from $\mathcal{N}(\mu_h,0.1)$. }
\label{pbw-demo}
\end{figure}

\paragraph{Parametric BipedalWalker env.} We test our modified AGAIN variants along with baselines on an existing parametric BipedalWalker environment proposed in \cite{portelas2019}, which generates walking tracks paved with stumps whose height and spacing are defined by a $2$D parameter vector used for the procedural generation of tasks. We keep the original bounds of this task space, i.e. we bound the stump-height dimension to $\mu_h \in [0,3]$ and the stump-spacing dimension to $\delta_s \in [0,6]$. As in their work, we also test our teachers when the learning agent is embodied in a modified short-legged walker, which constitutes an even more challenging scenario (as the task space is unchanged, i.e. more unfeasible tasks). The agent is rewarded for keeping its head straight and going forward and is penalized for torque usage. The episode is terminated after 1) reaching the end of the track, 2) reaching a maximal number of $2000$ steps, or 3) head collision (for which the agent receives a strong penalty). See figure \ref{pbw-demo} for visualizations.

\paragraph{Results} To perform our experiments, we ran each condition for either $10$Millions (IN and AGAIN variants) or $20$Millions (others)  environment steps ($30$ repeats). The preliminary ALP-GMM runs used in IN and AGAIN variants correspond to the first $10$ Million steps of the ALP-GMM condition (whose end-performance after $20$ Million steps is reported in table \ref{results-table}. All teacher variants are tested when paired with a Soft-Actor Critic \cite{sac} student, with same hyperparameters as in the Parkour experiments (see app. \ref{ann:parkour}). Performance is measured by tracking the percentage of mastered tasks (i.e. $r>230$) from a fixed test set of $100$ tasks sampled uniformly over the task space. We thereafter report results for $2$ independent experiments done with either default walkers or short walkers.

\begin{figure}
\centering\includegraphics[width=0.5\columnwidth]{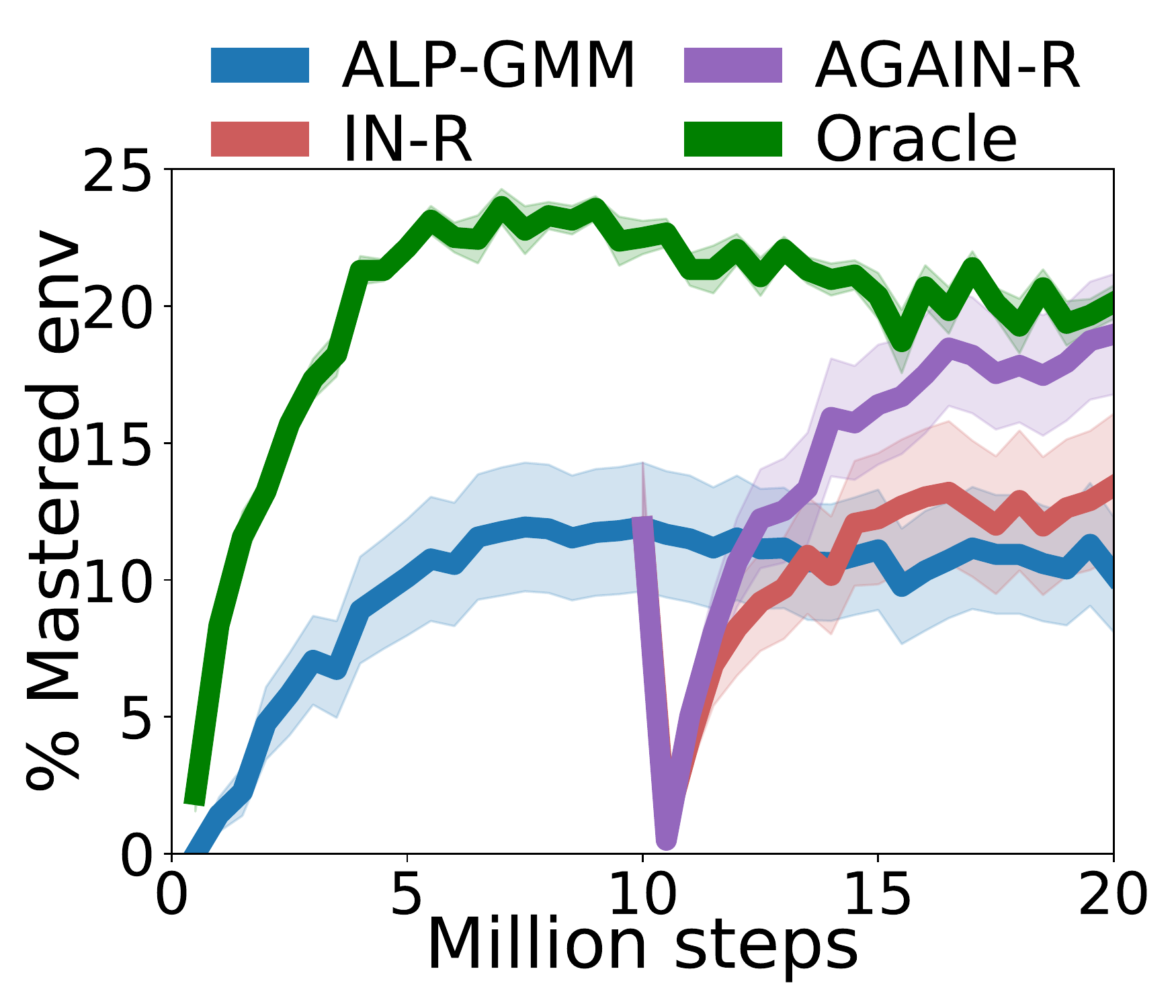}
\caption{\footnotesize{Given a single DRL student to train, AGAIN outperforms ALP-GMM in a parametric BipedalWalker environment. sem plotted, 30 seeds.}}\label{trying-again-compact-app}
\end{figure} 
\textit{Is re-training from scratch beneficial? - } The end performances of all tested conditions are summarized in table \ref{results-table}. Interestingly, retraining the DRL agent from scratch in the second run gave superior end performances than fine-tuning using the weights of the first run \textit{in all tested variants}. This showcases the brittleness of gradient-based training and the difficulty of transfer learning. Despite this, even fine-tuned variants reached superior end-performances than classical ALP-GMM, meaning that the change in curriculum strategy in itself is already beneficial.

\textit{Is it useful to re-use ALP-GMM in the second run? - } In the default walker experiments, AGAIN-R, T and P conditions mixing ALP-GMM and IN in the second run reached lower mean performances than their respective IN variants. However, the exact opposite is observed for IN-R and IN-T variants in the short walker experiments. This can be explained by the difficulty of short walker experiments for ACL approaches, leading to $16/30$ preliminary 10M steps long ALP-GMM runs to have a mean end-performance of $0$, compared to $0/30$ in the default walker experiments. All these run failures led to many GMMs lists $\mathcal{C}$ used in IN to be of very low-quality, which illustrates the advantage of AGAIN that is able to emancipate from IN using ALP-GMM.

\textit{Highest-performing variants. - } Consistently with the precedent analysis, mixing ALP-GMM with IN in the second run is not essential in default walker experiments, as the best performing ACL approach is IN-P. This most likely suggests that the improved adaptability of the curriculum when using AGAIN is outbalanced by the added noise (due to the low task-exploration). However in the more complex short walker experiments, mixing ALP-GMM with IN is essential, especially for AGAIN-R, which substantially outperforms ALP-GMM and other AGAIN and IN variants (see fig. \ref{trying-again-compact}), reaching a mean end performance of $19.0$. The difference in end-performance between AGAIN-R and Oracle, our hand-made expert using privileged information who obtained $20.1$, is not statistically significant ($p=0.6$).

\begin{table*}[]
\caption{\footnotesize{\textbf{Experiments on Parametric BipedalWalker} The avg. perf. with std. deviation after 10 Millions steps (IN and AGAIN variants) or 20 Million steps (others) is reported (30 seeds). For IN and AGAIN we also test variants that do not retrain the weights of the policy used in the second run \textit{from scratch} but rather \textit{fine-tune} them from the preliminary run.$\mathbf{^{*/-}}$ Indicates whether perf. difference with ALP-GMM is statistically significant ie. $p<0.05$ in a post-training Welch's student t-test ($\mathbf{^{*}}$ for performance advantage w.r.t ALP-GMM and $\mathbf{^{-}}$ for perf. disadvantage).}}
\vspace{0.3cm}
    \footnotesize
    \centering
\begin{tabular}{@{}lll@{}}
\toprule
Condition            & Short walker & Default walker \\ \midrule
AGAIN-R               & $19.0\pm12.0^*$               & $41.6\pm6.3^*$           \\
AGAIN-R(fine-tune)    & $11.4\pm12.9$                 & $39.9\pm4.6$           \\
IN-R                  & $13.4\pm14.4$                 & $43.5\pm9.6^*$           \\
IN-R(fine-tune)       & $11.2\pm12.3$                 & $40.8\pm5.6$           \\
AGAIN-T               & $15.1\pm11.9$                 & $40.6\pm11.5$           \\
AGAIN-T(fine-tune)    & $11.4\pm11.8$                 & $40.6\pm3.8^*$           \\
IN-T                  & $13.5\pm13.3$                 & $43.5\pm6.1^*$           \\
IN-T(fine-tune)       & $10.7\pm12.3$                 & $40.3\pm7.6$           \\
AGAIN-P               & $13.6\pm12.5$                 & $41.9\pm5.1^*$           \\
AGAIN-P(fine-tune)    & $11.1\pm12.0$                 & $41.5\pm3.9^*$           \\
IN-P                  & $14.5\pm12.6$                 & $\mathbf{44.3}\pm3.5^*$  \\
IN-P(fine-tune)       & $12.2\pm12.5$                 & $41.1\pm3.8^*$           \\
ALP-GMM               & $10.2\pm11.5$                 & $38.6\pm3.5$           \\
Oracle                & $\mathbf{20.1}\pm3.4^*$         & $27.2\pm15.2^-$           \\
Random                & $2.5\pm5.9^-$                   & $20.9\pm11.0^-$           \\ \bottomrule
\end{tabular}
    \label{results-table}
\end{table*}

\begin{figure*}[htb!]
\centering
\subfloat[\textbf{Pool-based} IN]{\includegraphics[width=0.31\textwidth]{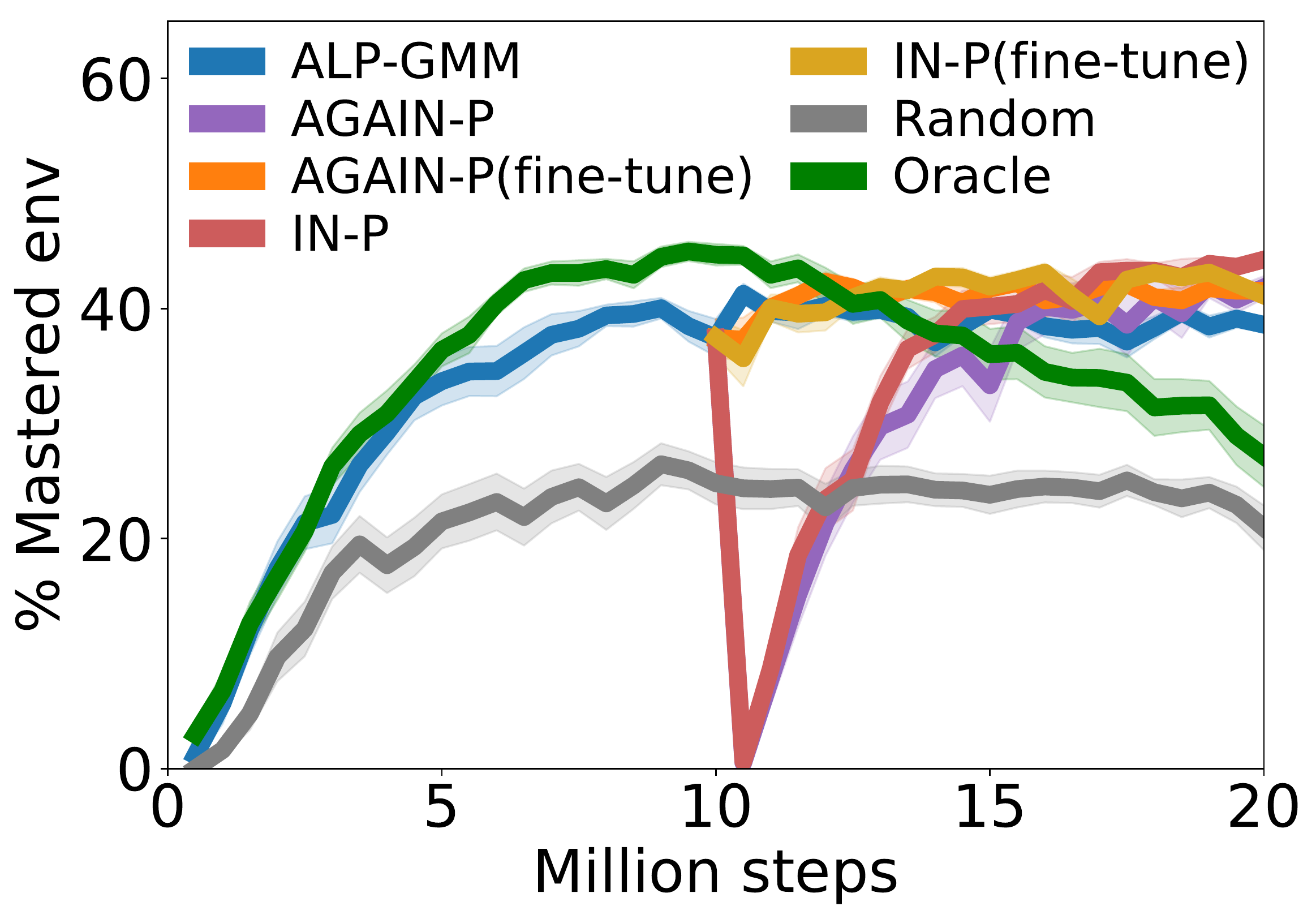}}
\subfloat[\textbf{Time-based} IN]{\includegraphics[width=0.31\textwidth]{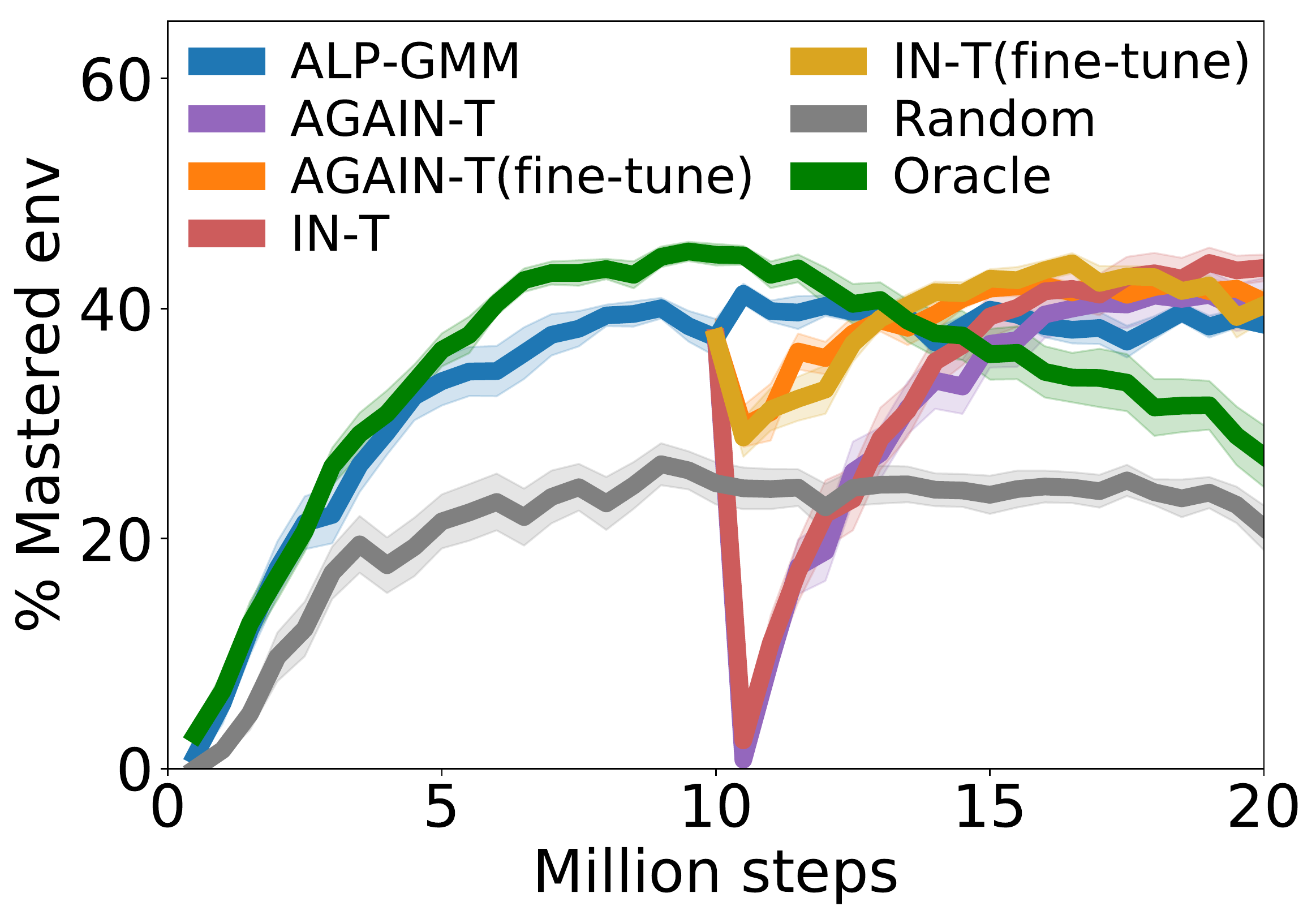}}
\subfloat[\textbf{Reward-based} IN]{\includegraphics[width=0.31\textwidth]{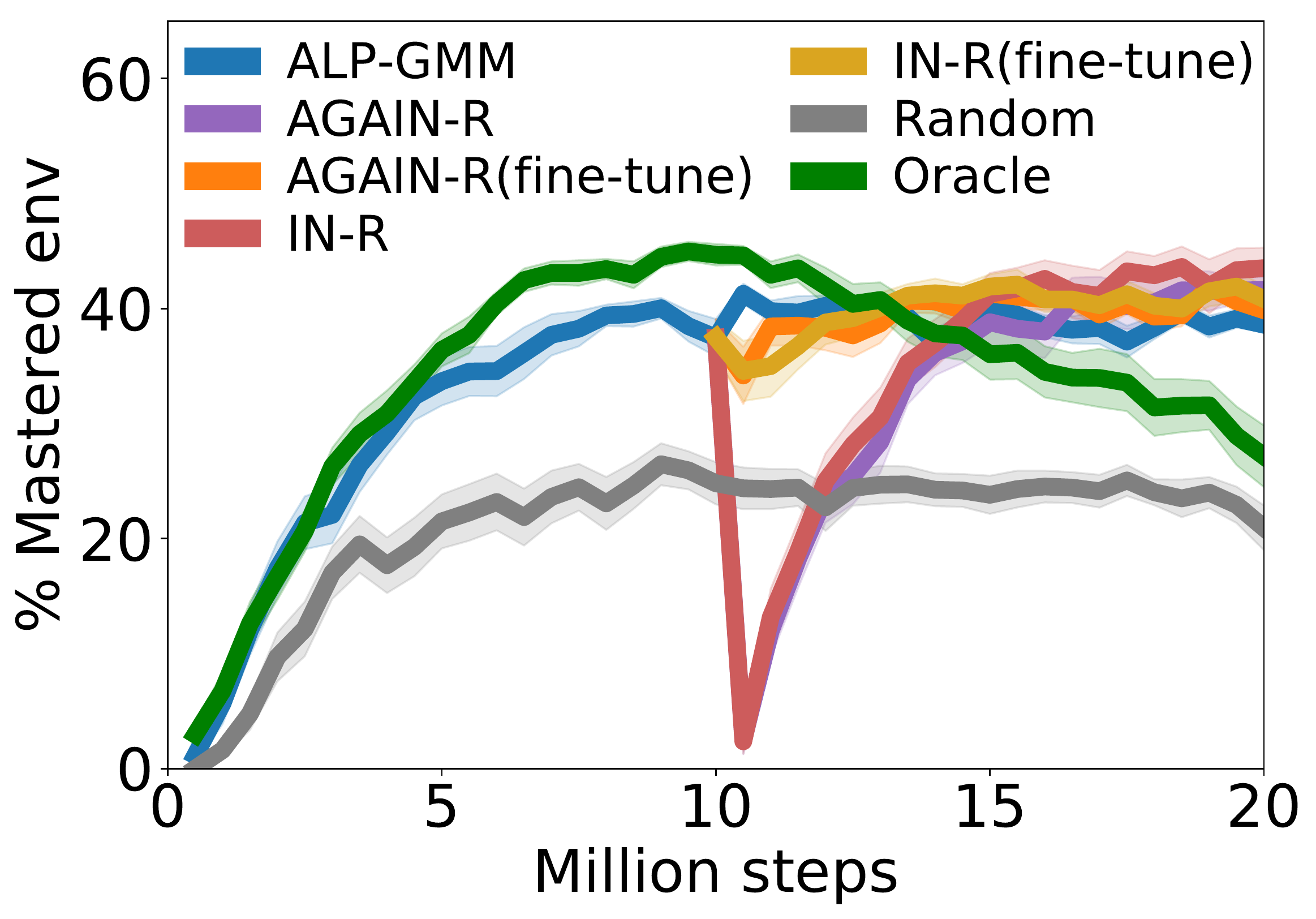}}
\caption{\footnotesize{\textbf{Evolution of performance across 20M environment steps of each condition with default bipedal walker.} Each point in each curve corresponds to the mean performance (30 seeds), defined as the percentage of mastered tracks (ie. $r>230$) on a fixed test set. Shaded areas represent the standard error of the mean. Consistently with \cite{portelas2019}, which implements a similar approach, Oracle is prone to forgetting with default walkers due to the strong shift in task subspace (which is why it is not the best performing condition for default walker experiments.}}
\label{default-exps-curves}
\end{figure*}

\begin{figure*}[htb!]
\centering
\subfloat[\textbf{Pool-based} IN]{\includegraphics[width=0.31\textwidth]{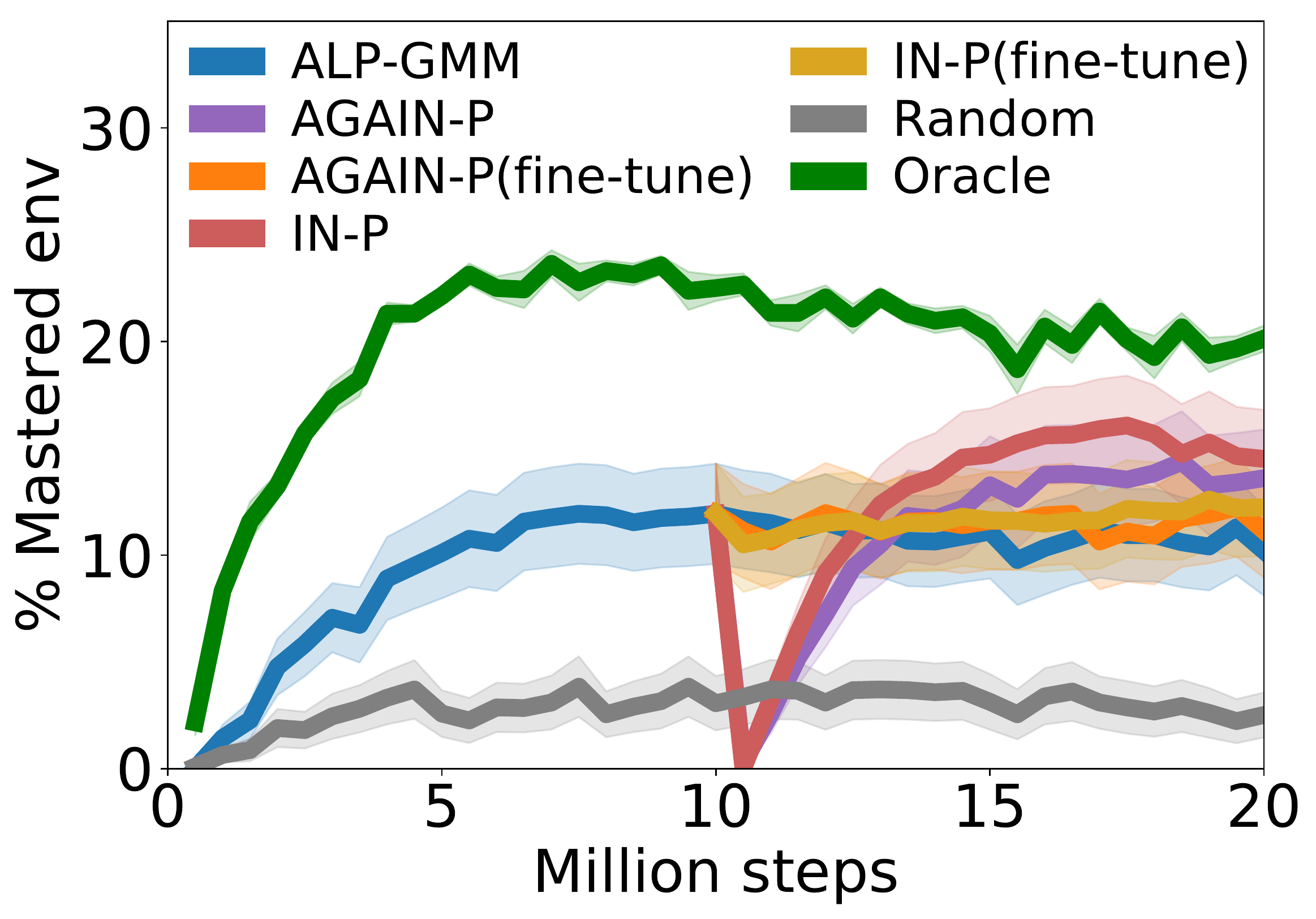}}
\subfloat[\textbf{Time-based} IN]{\includegraphics[width=0.31\textwidth]{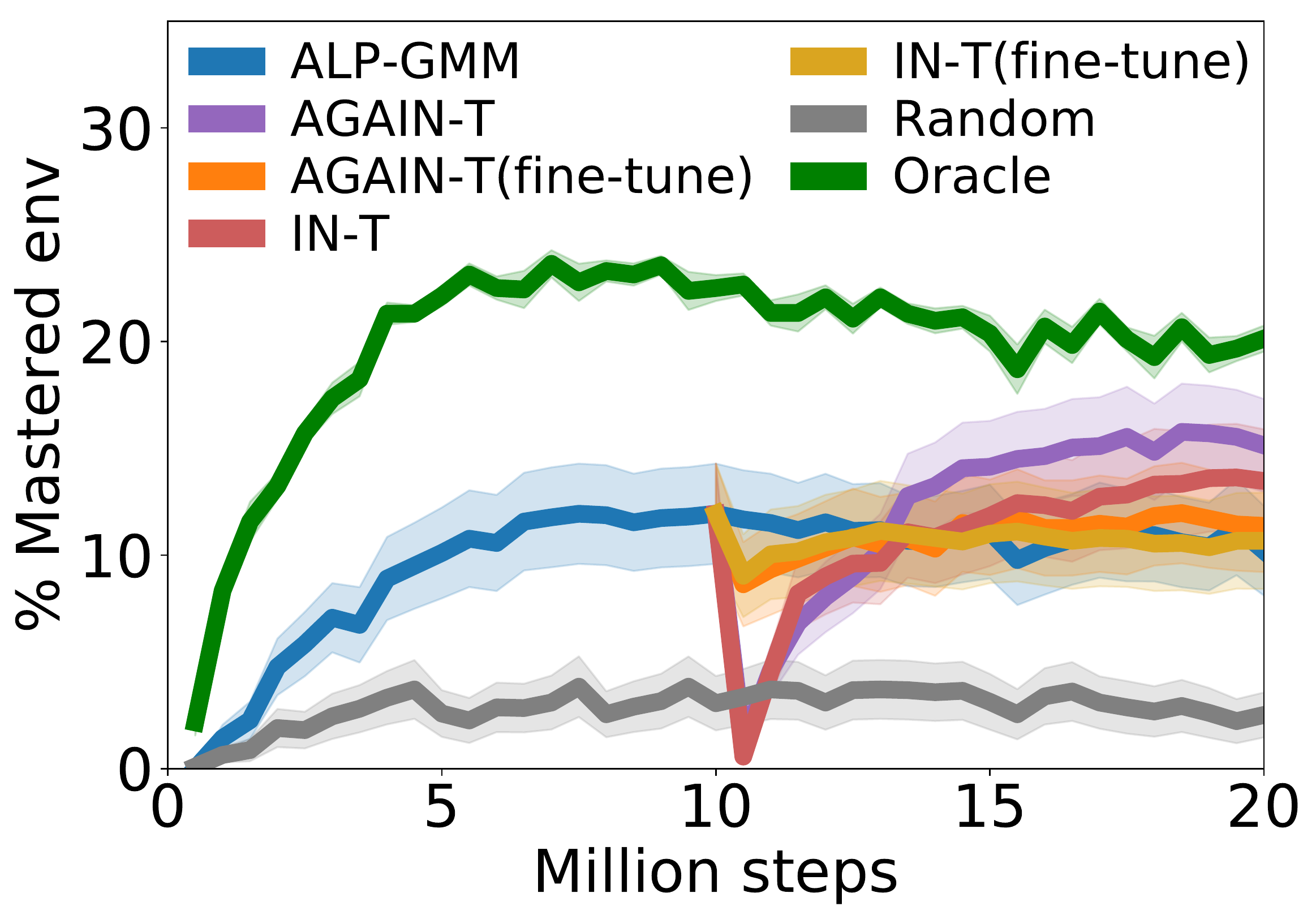}}
\subfloat[\textbf{Reward-based} IN]{\includegraphics[width=0.31\textwidth]{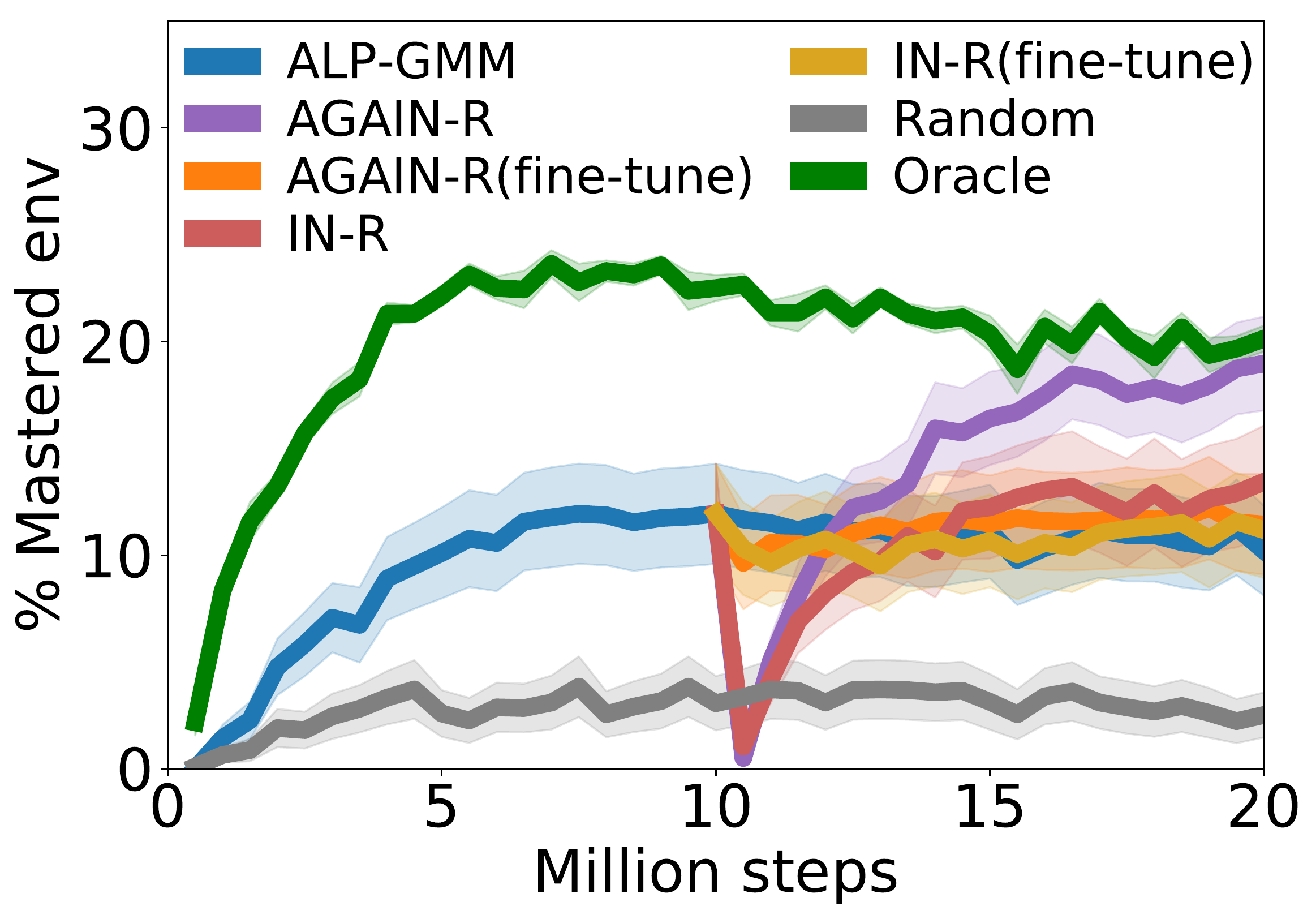}}
\caption{\footnotesize{\textbf{Evolution of performance across 20M environment steps of each condition with short bipedal walker.} Each point in each curve corresponds to the mean performance (30 seeds), defined as the percentage of mastered tracks (ie. $r>230$) on a fixed test set. Shaded areas represent the standard error of the mean.}}
\label{short-exps-curves}
\end{figure*}
\end{document}